\documentclass[sigconf]{acmart}
\usepackage[ruled,vlined,linesnumbered,noend]{algorithm2e} 
\usepackage[capitalise]{cleveref} 
\usepackage{bm} 
\crefname{equation}{Eq.}{Eqs.} 
\crefname{section}{Sec.}{Secs.} 
\usepackage{xcolor} 
\DeclareMathOperator*{\argmin}{arg\,min} 
\DeclareMathOperator*{\argmax}{arg\,max} 
\usepackage{multirow} 
\usepackage{makecell} 
\usepackage{colortbl} 
\usepackage{placeins}
\crefname{figure}{Fig.}{Figs.}
\crefname{table}{Tab.}{Tabs.}
\crefname{section}{Sec.}{Secs.}

\AtBeginDocument{%
  }

\setcopyright{acmlicensed}
\copyrightyear{2026}
\acmYear{2026}
\setcopyright{cc}
\setcctype{by}
\acmConference[MM '26]{Proceedings of the 34th ACM International Conference on Multimedia}{November 10--14, 2026}{Rio de Janeiro, Brazil}
\acmBooktitle{Proceedings of the 34th ACM International Conference on Multimedia (MM '26), November 10--14, 2026, Rio de Janeiro, Brazil}
\acmDOI{10.1145/3767308.3836526}
\acmISBN{979-8-4007-2213-4/2026/11}

\makeatletter
\newcommand{\correspondingauthornote}{%
  \g@addto@macro\@authornotes{%
    \begingroup
    \renewcommand{\thefootnote}{\ding{41}}%
    \footnotetext{Corresponding authors.}%
    \endgroup}}
\makeatother

\begin{document}

\title[Probe and Skip: Self-Predictive Token Skipping for Efficient Long-Context LLM Inference]{Probe and Skip: Self-Predictive Token Skipping \\ for Efficient Long-Context LLM Inference}

\author{Zimeng Wu}
\authornote{Both authors contributed equally to this research.}
\email{zimengwu@buaa.edu.cn}
\orcid{0009-0005-2653-1402}
\affiliation{%
  \department{State Key Laboratory of Virtual Reality Technology and Systems}
  \institution{Beihang University}
  \city{Beijing}
  \country{China}
}

\author{Donghao Wang}
\authornotemark[1]
\email{wonderhow@buaa.edu.cn}
\orcid{0009-0007-9393-5532}
\affiliation{%
  \department{State Key Laboratory of Virtual Reality Technology and Systems}
  \institution{Beihang University}
  \city{Beijing}
  \country{China}
}

\author{Chaozhe Jin}
\email{chaozhejin@buaa.edu.cn}
\orcid{0009-0009-9305-3884}
\affiliation{%
  \department{State Key Laboratory of Virtual Reality Technology and Systems}
  \institution{Beihang University}
  \city{Beijing}
  \country{China}
}

\author{Jiaxin Chen}
\correspondingauthor
\email{jiaxinchen@buaa.edu.cn}
\orcid{0000-0002-0112-4166}
\affiliation{%
  \department{State Key Laboratory of Virtual Reality Technology and Systems}
  \institution{Beihang University}
  \city{Beijing}
  \country{China}
}

\author{Yunhong Wang}
\correspondingauthor
\email{yhwang@buaa.edu.cn}
\orcid{0000-0001-8001-2703}
\affiliation{%
  \department{State Key Laboratory of Virtual Reality Technology and Systems}
  \institution{Beihang University}
  \city{Beijing}
  \country{China}
}

\renewcommand{\shortauthors}{Zimeng Wu, Donghao Wang, Chaozhe Jin, Jiaxin Chen, and Yunhong Wang}

\begin{abstract}
  Long-context inference enhances the reasoning capability of Large Language Models (LLMs), but incurs significant computational overhead. Token-oriented methods, such as pruning and skipping, have shown great promise in reducing inference latency, yet still suffer from inherently insufficient structure optimization, outdated selection criteria, and redundancy interference, resulting in suboptimal speed-accuracy trade-off. To address these issues, we propose a novel training-free framework dubbed Self-Predictive Token Skipping (SPTS), for efficient long-context LLM inference. Specifically, motivated by probing the influence of target layers prior to skipping, we design two selective token skipping strategies for typical structures, including Partial Attention Probing (PAP) for multi-head attention and Low-rank Transformation Probing (LTP) for feed forward network. The former selects informative tokens via partial forward attention computation, while the latter constructs a low-rank proxy network to predict token transformations. In addition, a Multi-Stage Delayed Pruning (MSDP) strategy reallocates skipping budgets and progressively removes redundant tokens across layers. Extensive experiments display the effectiveness of our method, achieving up to 2.46$\times$ and 2.29$\times$ speedups for prefilling and end-to-end generation, respectively, while maintaining state-of-the-art accuracy. The source code is available at \url{https://github.com/Wuzimeng/SPTS}.
\end{abstract}

\ccsdesc[500]{Computing methodologies~Natural language processing}
\ccsdesc[300]{Computing methodologies~Artificial intelligence}
\ccsdesc[100]{Computing methodologies~Computer vision}

\keywords{Token Pruning, Efficient Inference, Large Language Model}

\correspondingauthornote

\maketitle

\begin{figure}[!t]
  \begin{center}
    \centerline{\includegraphics[width=\linewidth]{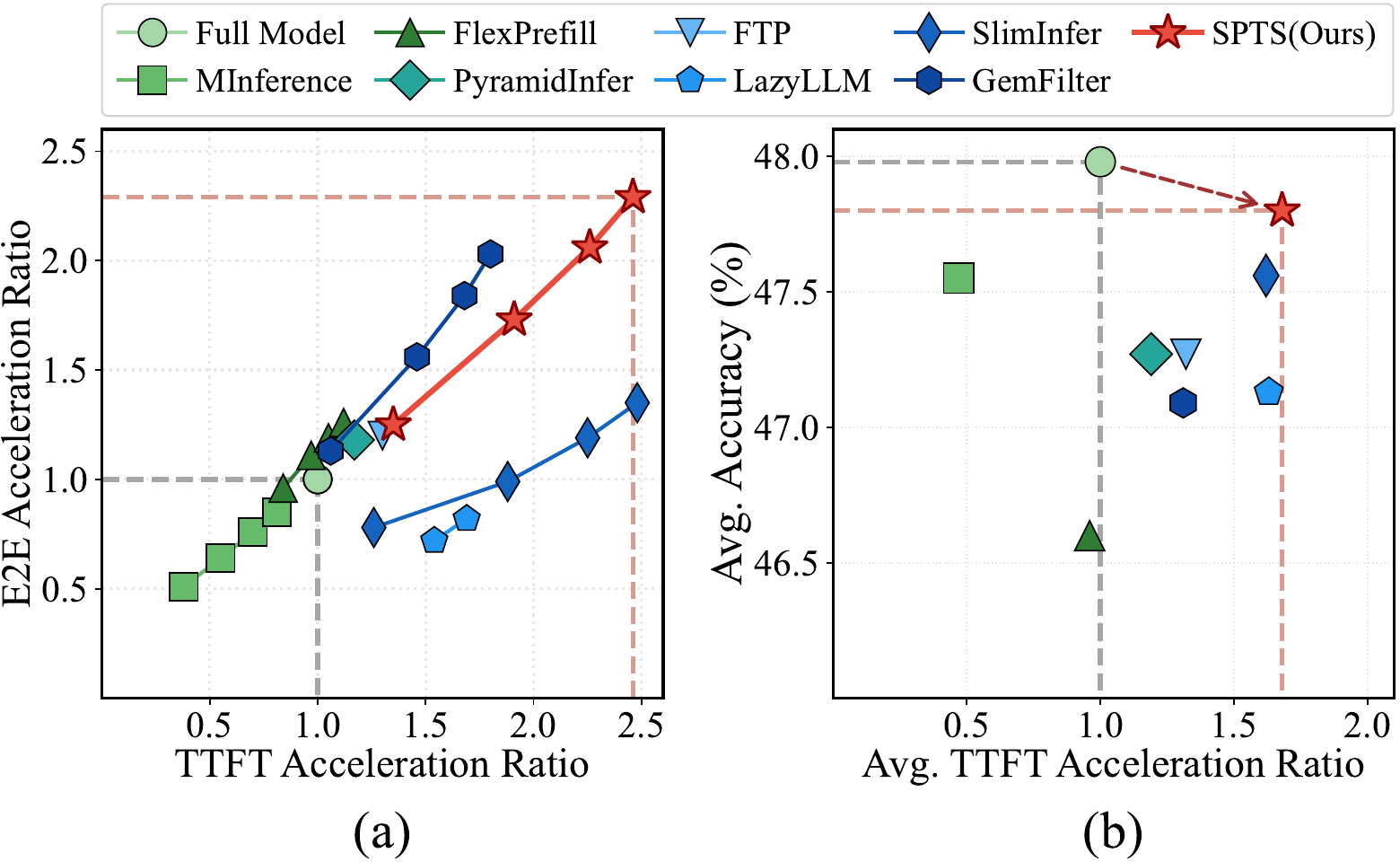}}
    \vspace{-8pt}
    \caption{
      Efficiency-accuracy comparison on LLaMA-3.1-8B-Instruct \cite{llama8b}. (a) SPTS accelerates both prefilling and decoding, and achieves consistent speedups in first token (TTFT) and end-to-end (E2E) generation. (b) SPTS maintains higher accuracy at a more aggressive TTFT speedup ratio, with metrics averaged over the LongBench \cite{longbench} dataset.
    }
    \label{fig:tradeoff}
  \end{center}
  \vspace{-15pt}
\end{figure}
\section{Introduction}

Recent advances in Large Language Models (LLMs) have significantly improved their capabilities in reasoning over long contexts \cite{techniques_to_extend_long_context,seagent,curriculumrlaif}, enabling a wide range of applications such as multi-document question answering \cite{hotpotqa}, code completion \cite{repobench}, and even broader multimodal tasks \cite{wowseg,crystal,unichange}.
However, as the input sequence length increases, the computational overhead grows sharply, limiting the inference speed in both the prefilling and decoding phases under the standard auto-regressive paradigm \cite{lclm_survey}.

In order to reduce inference latency, a variety of acceleration strategies have been explored \cite{llmpruner, pyramidinfer, flexprefill, samp}. Token pruning provides an effective solution by directly removing redundant tokens \cite{pyramidinfer}, but incurs an inflexible trade-off between end-to-end acceleration and preservation of critical information \cite{ftp}. The time cost of recomputing the pruned tokens typically outweighs the speed gains from aggressive compression \cite{lazyllm, sliminfer}. In contrast, sparse attention \cite{flexprefill} and FFN skipping \cite{ftp} select a subset of tokens to involve in attention or FFN computation and ignore the rest, decreasing latency while maintaining accuracy without dropping context tokens in forward pass. Nevertheless, they still suffer from three key limitations: 
1) \emph{Insufficient structure optimization}. Existing methods are typically designed for either Multi-Head Attention (MHA) or Feed Forward Network (FFN) modules, where pursuing high acceleration ratios induces
severe performance degradation in the optimized module. Meanwhile, the unoptimized counterpart becomes a bottleneck, ultimately hindering the overall speedup.
2) \emph{Suboptimal token selection by outdated criteria}. To select crucial tokens for computation, most existing methods rely on outdated criteria such as attention score from preceding layers \cite{ftp}, which fail to reflect importance in the layer to be computed, thereby leading to suboptimal token skipping decisions.
3) \emph{Over preservation of tokens introducing interference}. Retaining full context throughout deep layers is often unnecessary, as the redundant tokens require repeated evaluations for selection and diminish the discriminability between informative and uninformative tokens, thereby leading to biased selection.

To address these limitations, we propose SPTS, a training-free framework for efficient long-context LLM inference. Specifically, we first analyze the feasibility of token skipping across deep layers and emphasize our main principle of self-predictive token skipping: 
\textbf{A token’s eligibility for skipping is determined by its potential impact within the target layer}. 
Based on this insight, we design specific criteria to select active tokens for typical structures. As for MHA, we employ Partial Attention Probing (PAP) that estimates token contribution by probing a small portion of attention score. As for FFN, we propose Low-rank Transformation Probing (LTP) that pre-constructs a low-rank proxy network to approximate the original dense FFN, guiding token selection via proxy-induced variations combined with attention scores.
To further improve efficiency and mitigate redundancy interference, a Multi-Stage Delayed Pruning (MSDP) strategy is developed, which divides layers into stages and applies pruning at stage boundaries after token skipping, thereby progressively reducing both active and candidate tokens.

In summary, our main contributions lie in three-fold:

1) We propose SPTS, a novel training-free framework for accelerating long-context LLM inference, which achieves remarkable speedups in both prefilling and decoding phases.

2) We design Partial Attention Probing (PAP) and Low-rank Transformation Probing (LTP) for MHA and FFN modules, respectively, by applying self-predictive information to boost token selection. We also develop a Multi-Stage Delayed Pruning (MSDP) strategy to suppress the redundancy-induced interference.

3) We conduct extensive experiments on representative LLMs, showing that SPTS outperforms state-of-the-art approaches in both efficiency and accuracy. As shown in \cref{fig:tradeoff}, our method achieves up to 2.46$\times$ and 2.29$\times$ speedups on LLaMA-3.1-8B-Instruct \cite{llama8b} for prefilling and end-to-end generation, respectively, with minimal performance degradation. Additionally, SPTS also excels on Multimodal Large Language Models (MLLMs), exhibiting generalization capability across both long textual and visual contexts.

\section{Related Work}
\subsection{LLM Inference and Efficiency}
Prevalent LLM inference involves two phases: prefilling and auto-regressive decoding \cite{llm_survey}. During prefilling, the model processes the input sequence to generate the first token, computing full causal attention and storing the key and value representations in the Key-Value (KV) Cache. In decoding, each newly generated token is fed back as input and attends to the KV Cache to produce the next token. For long-context tasks with thousands of input tokens \cite{longbench}, both phases pose significant challenges in inference efficiency \cite{resource_efficient_llm_survey,resilphase}. 
A similar issue also arises in prevalent MLLMs \cite{li2026evolving,fleg}, as long visual token sequences are concatenated and fed into the LLM \cite{llavaimproved, llavanext}.

FlashAttention \cite{flashattention2} is a key baseline for efficient LLM deployment, accelerating MHA via optimized memory access. Sparse attention further reduces attention computation over long context \cite{flexprefill, minference}, but both schemes leave the computation-heavy FFN largely unoptimized. Token-level methods, such as token pruning \cite{streamllm} and skipping \cite{ftp}, offer complementary acceleration and can be combined with FlashAttention for more comprehensive speedup.

\begin{figure*}[htbp]
  \begin{center}
    \centerline{\includegraphics[width=0.95\textwidth]{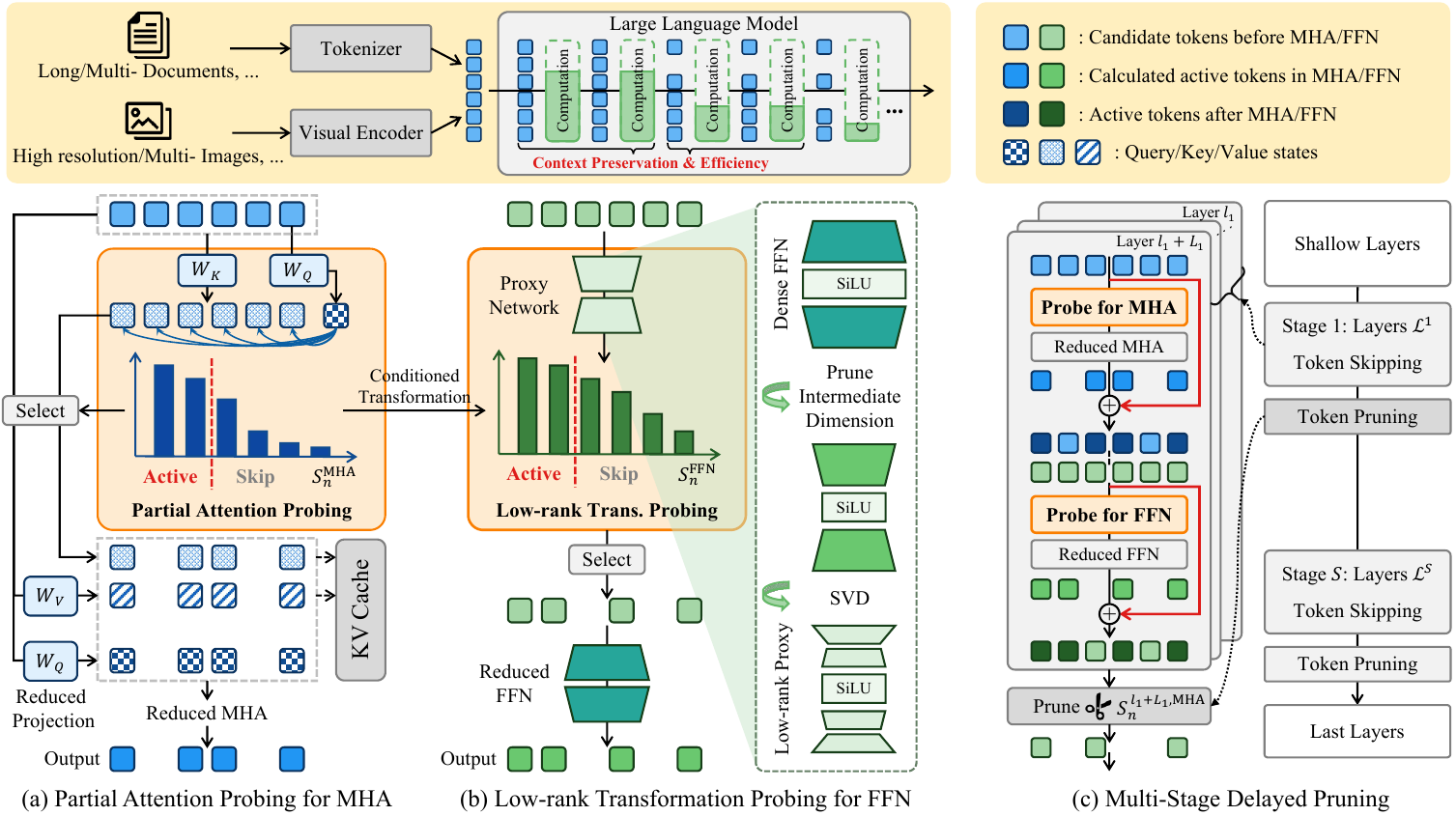}}
    \vspace{-10pt}
    \caption{
      Framework overview of SPTS. (Top) The overall data flow with SPTS. SPTS accelerates the LLM inference over both long textual and visual inputs by token skipping, reducing computation cost while preserving contextual information. (Bottom) Illustration of the main modules of SPTS. (a) Partial Attention Probing (PAP) performs token skipping in MHA by pre-computing a small fraction of attention to identify informative tokens, which then participate in reduced attention computation. (b) Low-rank Transformation Probing (LTP) enables token skipping in FFN by leveraging an offline-derived low-rank proxy network to probe token transformations, which are combined with attention-based information to guide token selection for reduced feed-forward. (c) Multi-Stage Delayed Pruning (MSDP) partitions the network into multiple stages, where the set of candidate tokens remains fixed within each stage and is reduced through pruning at stage boundaries.
    }
    \label{fig:framework}
  \end{center}
  \vspace{-18pt}
\end{figure*}

\subsection{Token Pruning}
Token pruning reduces computation by shortening the sequence \cite{comp,sremerge}, with effectiveness depending on when it is applied. Prior works commonly focus on the decoding phase (also known as KV Cache eviction) \cite{snapkv, chunkkv,cake,calidrop}, yet at the cost of increased prefilling latency. Recognizing prefilling as a bottleneck, recent works shift pruning to this phase and introduce strategies to compensate for information loss. PyramidInfer \cite{pyramidinfer} and GemFilter \cite{gemfilter} enhance token selection by lengthening the pruning window and using extra forward passes, respectively, which adds non-negligible overhead. LazyLLM \cite{lazyllm} and SlimInfer \cite{sliminfer} recompute the pruned tokens during decoding, leading to substantial generation slowdown. Overall, token pruning methods face an inherent trade-off between effective acceleration and faithful information preservation.

In MLLMs, most methods perform token reduction during prefilling \cite{fastv, sparsevlm, pyramiddrop, llavaprumerge}, whereas our method provides a context-preserving acceleration scheme prior to pruning.

\subsection{Token Skipping}

Token skipping leverages residual connections in modern network architectures to route a subset of tokens through the shortcut paths only, thereby reducing computation while preserving sequence integrity. Early methods either skip all tokens in fixed layers \cite{shortgpt} or employ learned predictors to determine which tokens to bypass \cite{ftp_route}. FTP \cite{ftp} explores a training-free token skipping for FFN acceleration, but encounters limited speedup. Moreover, its attention-based criterion leads to suboptimal token selection.

In this work, we further explore training-free token skipping strategies tailored for long-context LLMs, aiming to achieve more comprehensive and accurate acceleration.

\section{Methodology}

\subsection{Framework Overview}
Most prevalent LLMs adopt the standard Transformer \cite{transformer} architecture, where stacked MHA and FFN blocks dominate computation. Therefore, we target both to enable more aggressive acceleration. 

Notably, the two blocks share the same residual structure with shortcut connections. Without loss of generality, we denote $\mathcal{F}(\cdot)\in \{\mathcal{F}_{\mathrm{MHA}}(\cdot), \mathcal{F}_{\mathrm{FFN}}(\cdot)\}$ as the internal computation of either MHA or FFN, and let $\bm{X},~\bm{Y}\in \mathbb{R}^{N\times D}$ represent the input and output hidden states of a block, with $N$ the sequence length and $D$ the feature dimension. The block-wise operation can be formally expressed as:
\begin{equation}
    \bm{Y} = \bm{X} + \mathcal{F}(\bm{X}).
    \label{eq:residual}
\end{equation}

\begin{figure*}[htbp]
  \begin{center}
    \centerline{\includegraphics[width=\textwidth]{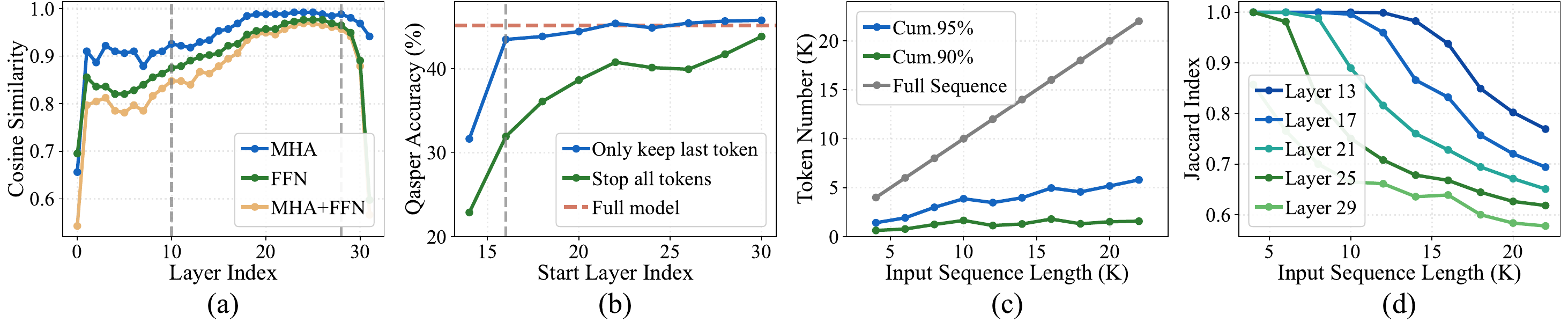}}
    \vspace{-10pt}
    \caption{Empirical observations on LLaMA-3.1-8B-Instruct. (a) Layer-wise average cosine similarity of tokens before and after MHA, FFN, and the full Transformer block (MHA+FFN). (b) Accuracy fluctuations induced by disabling MHA updates for all input tokens, or for all of them except the last, starting from a given layer. (c) Number of tokens contributing to 90\% and 95\% of cumulative attention across deep layers for input sequences of varying lengths. (d) Jaccard index between consecutive-layer token sets selected via the attention criterion under a progressively reduced token budget, across distinct input lengths.}
    \label{fig:motivation}
  \end{center}
  \vspace{-18pt}
\end{figure*}

Empirically, as shown in \cref{fig:motivation}(a), $\bm{X}$ and $\bm{Y}$ exhibits increasing similarity in deep layers across both blocks, indicating that only minor transformations are applied. This suggests that the residual shortcut preserves most contextual information, thereby motivating skipping $\mathcal{F}(\cdot)$ for a subset of tokens in each block.

Concretely, let $\mathcal{T}\!=\!\{1,2,\ldots,N\}$ denote the indices along the sequence dimension of $\bm{X}$, representing the candidate tokens. A subset $\mathcal{T}_{\mathrm{active}}\subset \mathcal{T}$ with cardinality $|\mathcal{T}_{\mathrm{active}}|=M$ is selected to identify active tokens. Then, the corresponding hidden states form a reduced matrix $\hat{\bm{X}}=\bm{X}[\mathcal{T}_{\mathrm{active}},~:]\in \mathbb{R}^{M\times D}$, which is processed exclusively by $\mathcal{F}(\cdot)$, while the remaining tokens follow an identity mapping, as formalized in \cref{eq:skip} and illustrated in~\cref{fig:framework}(c).
\begin{equation}
    \begin{split}
    \hat{\bm{Y}}[\mathcal{T}_{\mathrm{active}},~:] &= \hat{\bm{X}} + \mathcal{F}(\hat{\bm{X}}), \\
    \hat{\bm{Y}}[\mathcal{T}/\mathcal{T}_{\mathrm{active}},~:] &= \bm{X}[\mathcal{T}/\mathcal{T}_{\mathrm{active}},~:]. \\
    \end{split}
    \label{eq:skip}
\end{equation}

Building on this, our SPTS operates in the prefilling phase to enable inference-time acceleration. As illustrated in \cref{fig:framework}, full computation is retained in shallow layers to preserve complex information processing, while token skipping is applied in deeper layers. To address the aforementioned challenges, we introduce PAP and LTP for self-predictive selection of $\mathcal{T}_{\mathrm{active}}$, and design the MSDP strategy to regulate the overall token flow. Notably, SPTS also compresses the KV Cache during MHA skipping, enabling speedup in both prefilling and decoding phases at once. We provide detailed pseudocode of the inference procedure in the \emph{Appendix}.

\subsection{Partial Attention Probing for MHA}
\label{sec:pap}

Empirically, MHA blocks propagate information via causal attention, and prior studies suggest that most information diffusion occurs in shallow layers \cite{sliminfer}, making them sensitive to computation reduction. We thus analyze the role of deep-layer MHA to facilitate its acceleration. As shown in \cref{fig:motivation}(b), halting updates for almost all context tokens in deep MHA has little impact, whereas stopping the last-token update significantly degrates performance. This indicates that deep MHA primarily aggregates preceding information into the last token, supporting subsequent generation. The observation is also consistent with the nearby-focusing effect of RoPE \cite{rope_analysis}, under which skipping-induced errors decay with positional distance, making the last token, which is closest to the generation position, particularly important.

Accordingly, under a fixed computational budget, MHA token skipping can be formulated as the optimization problem in \cref{eq:mha_optimization}, aiming to minimize information loss in the last token.
\begin{equation}
    \hat{\mathcal{T}}^{\mathrm{MHA}}_{\mathrm{active}} = \argmin_{\mathcal{T}_{\mathrm{active}}\subset \mathcal{T}, |\mathcal{T}_{\mathrm{active}}|=M} \|\hat{\bm{Y}}_{N}-\bm{Y}_N\|_2.
    \label{eq:mha_optimization}
\end{equation}

However, this problem is NP-hard, rendering exact solutions infeasible within a single forward pass. Noting that attention mechanisms explicitly capture the information flow, we probe a lightweight, attention-based signal as a practical surrogate to approximate each token’s contribution and guide the skipping process.

Specifically, as illustrated in~\cref{fig:framework}(a), before the target $\mathcal{F}_{\mathrm{MHA}}(\cdot)$, we first compute the key states for all tokens as $\bm{K} = \bm{X}\bm{W}_K$, and the query state only for the last token as $\bm{q} = \bm{X}_N\bm{W}_Q$, where $\bm{W}_Q, \bm{W}_K$ are projection matrices. Then, the Softmax-normalized attention scores between $\bm{q}$ and $\bm{K}$ are calculated and averaged across all heads to derive a token-wise contribution score:
\begin{equation}
    S_n^{\mathrm{MHA}} = \frac{1}{H}\sum_{h=1}^H\mathrm{Softmax}(\frac{\bm{q}_h\bm{K}_h^T}{\sqrt{d_k}})[n],
    \label{eq:mha_importance}
\end{equation}
where $H$ and $d_k$ denote the number of attention heads and the dimensionality per head, respectively. Accordingly, the top-$M$ tokens are selected, yielding the active indices as:
\begin{equation}
    \mathcal{T}^{\mathrm{MHA}}_{\mathrm{active}}=\mathrm{TopK}(\{S_n^{\mathrm{MHA}}\}_{n=1}^{N}, M).
\end{equation}

Finally, for a computationally reduced $\mathcal{F}_{\mathrm{MHA}}(\cdot)$, query and value projections are computed only for the selected tokens $\hat{\bm{X}}$, while the corresponding key states are retrieved from the precomputed $\bm{K}$ as $\bm{K}[\mathcal{T}_{\mathrm{active}},~:]$. This further lowers the subsequent cost of attention and output projection.

Moreover, this MHA skipping strategy also enables on-the-fly KV Cache compression during prefilling. As \cref{fig:framework}(a) shows, only the key and value states of active tokens are cached, while skipped tokens still propagate to subsequent layers. 
Such selective caching reduces memory footprint and computational overhead in decoding, while providing more flexible KV Cache management that helps preserve inference accuracy, as further discussed in the \emph{Appendix}.

\subsection{Low-rank Transformation Probing for FFN}
\label{sec:ltp}

Unlike the inter-token propagation in MHA, FFN blocks perform independent intra-token representation refinement, formulated as:
\begin{equation}
    \mathcal{F}_{\mathrm{FFN}}(\bm{X}) = (\sigma(\bm{X}\bm{W}_{\mathrm{gate}})\odot \bm{X}\bm{W}_{\mathrm{up}})\bm{W}_{\mathrm{down}}^T,
    \label{eq:ffn_function}
\end{equation}
where $\bm{W}_{\mathrm{up}}, \bm{W}_{\mathrm{gate}}, \bm{W}_{\mathrm{down}}\in \mathbb{R}^{D\times D_{\mathrm{ff}}}$ are the projection matrices with intermediate dimension $D_{\mathrm{ff}}$, $\sigma$ is a non-linear activation (\emph{e.g.} SiLU), and $\odot$ is element-wise multiplication.

Accordingly, the objective of FFN token skipping is to minimize the representation discrepancy over the sequence. Given \cref{eq:residual} and the token-wise independence property, \emph{i.e.}, $\mathcal{F}_{\mathrm{FFN}}(\bm{X})[n]=\mathcal{F}_{\mathrm{FFN}}(\bm{X}_n)$, the objective can be derived as \cref{eq:ffn_optimization}. This indicates that the transformation magnitude of each token serve as an effective criterion, whereby tokens undergoing larger representation changes should be selected as the active subset.
\begin{equation}
    \begin{split}
        \hat{\mathcal{T}}^{\mathrm{FFN}}_{\mathrm{active}} & = \argmin_{\mathcal{T}_{\mathrm{active}}\subset \mathcal{T}, |\mathcal{T}_{\mathrm{active}}|=M} \|\hat{\bm{Y}}-\bm{Y}\|_2 \\
        & = \argmax_{\mathcal{T}_{\mathrm{active}}\subset \mathcal{T}, |\mathcal{T}_{\mathrm{active}}|=M} \sum_{n\in\mathcal{T}_{\mathrm{active}}}\|\mathcal{F}(\bm{X}_n)\|_2.
    \end{split}
    \label{eq:ffn_optimization}
\end{equation}

\textbf{Low-rank Proxy FFN.}
However, probing the exact solution of \cref{eq:ffn_optimization} during online inference is computationally prohibitive, as it contradicts the goal of reducing the cost of $\mathcal{F}_{\mathrm{FFN}}(\cdot)$ itself. Therefore, we adopt a lightweight proxy sub-network $f(\cdot)$ to estimate token-wise transformations. For accurate prediction, $f(\cdot)$ is explicitly designed to closely mimic the behavior of the original FFN:
\begin{equation}
    \hat{f}(\cdot) = \argmin_{f(\cdot)} \|\mathcal{F}_{\mathrm{FFN}}(\bm{X})-f(\bm{X})\|_2.
    \label{eq:proxy_network_optimization}
\end{equation}

Inspired by network pruning methods \cite{wanda,ukmp,miep}, we construct $f(\cdot)$ by initializing it from the original $\mathcal{F}_{\mathrm{FFN}}(\cdot)$ and slimming it through structured removal of projection matrix dimensions that have minimal impact, as illustrated in~\cref{fig:framework}(b).

Starting from \cref{eq:ffn_function}, we first reduce the intermediate dimension $D_{\mathrm{ff}}$ to $D_{\mathrm{low}}$ via a data-driven scheme. By forwarding a set of calibration text sequences through the model, we collect the hidden states immediately preceding $\mathcal{F}_{\mathrm{FFN}}(\cdot)$ to form a calibration set $\mathcal{G}$.
For each token $\bm{x}\in \mathcal{G}$, we compute its activation saliency vector as:
\begin{equation}
    z(\bm{x})=|\sigma(\bm{x}\bm{W}_{\mathrm{gate}})\odot \bm{x}\bm{W}_{\mathrm{up}}|\in \mathbb{R}^{D_{\mathrm{ff}}}.
\end{equation}
A channel-wise importance score for dimension $j$ is then estimated by averaging the top-$\rho$ fraction of its saliency values across $\mathcal{G}$:
\begin{equation}
    I_j = \mathrm{Mean}(\mathrm{TopK}(\{z(\bm{x})[j]\,|\,\bm{x}\in \mathcal{G}\}, \rho\times|\mathcal{G}|)).
    \label{eq:channel_importance}
\end{equation}
Intuitively, this criterion prioritizes intermediate dimensions consistently highly activated by the responsive tokens, thus preserving the most informative channels.
Accordingly, we select the top-$D_{\mathrm{low}}$ dimensions $\mathcal{C}\!=\!\mathrm{TopK}(\{I_j\}_{j=1}^{D_{\mathrm{ff}}}, D_{\mathrm{low}})$ and reduce the projection matrices as \cref{eq:channel_prune}. In the literature \cite{sparsegpt,wanda}, this calibration-based process remains training-free, as it requires no weight updates.
\begin{equation}
    \bm{W}'=\bm{W}[:,~\mathcal{C}],~\forall\bm{W}\in\{\bm{W}_{\mathrm{gate}},\bm{W}_{\mathrm{up}},\bm{W}_{\mathrm{down}}\}.
    \label{eq:channel_prune}
\end{equation}

However, dense matrix multiplications with $\bm{W}'$ remain costly. To further reduce computation, we apply a low-rank factorization by performing a Singular Value Decomposition (SVD) on $\bm{W}'$. Retaining the top-$r$ singular values yields a rank-$r$ approximation $\bm{W}'\approx \bm{U}\bm{V}$, where $\bm{U}\in \mathbb{R}^{D\times r}$ and $\bm{V}\in \mathbb{R}^{r\times D_{\mathrm{low}}}$ with $r\ll \min (D, D_{\mathrm{low}})$.

Finally, by integrating the two dimensionality reduction steps, the computation of proxy $f(\cdot)$ can be formalized as:
\begin{equation}
    \begin{split}
        f(\bm{X}) &= (\mathrm{Gate}(\bm{X})\odot\mathrm{Up}(\bm{X}))\bm{V}_{\mathrm{down}}^T\bm{U}_{\mathrm{down}}^T,~~~~~~ \\
        \mathrm{Gate}(\bm{X})&=\sigma(\bm{X}\bm{U}_{\mathrm{gate}}\bm{V}_{\mathrm{gate}}),~\mathrm{Up}(\bm{X})=\bm{X}\bm{U}_{\mathrm{up}}\bm{V}_{\mathrm{up}}.
    \end{split}
    \label{eq:proxy_ffn}
\end{equation}

\textbf{Conditioned Transformation for Token Selection.}
According to \cref{eq:ffn_optimization,eq:proxy_network_optimization}, the influence of an FFN block on token $\bm{X}_n$ can be estimated by the $l_2$-norm of its output from the proxy network:
\begin{equation}
    C^{\mathrm{FFN}}_n = \|f(\bm{X}_n)\|_2,
    \label{eq:ffn_transformation}
\end{equation}
which identifies tokens that are expected to undergo the largest representation transformations, and should therefore be prioritized for active computation rather than being skipped.

However, such token-independent estimation ignores task-related dependencies: some tokens with minor self-updates may still encode crucial information, and aggressively skipping them can introduce undesirable noise. To mitigate this, we skip tokens only when they are both minimally informative and exhibit limited changes, \emph{i.e.} with low transformation conditioned on a low information score.

Formally, we integrate the probed contribution score from \cref{eq:mha_importance} to capture task relevance. The two metrics are then multiplicatively fused to define a conditioned transformation score as \cref{eq:ffn_importance}, and the active set is ultimately determined by top-$M$ selection as \cref{eq:ffn_selection}.
\begin{equation}
    S^{\mathrm{FFN}}_n = C^{\mathrm{FFN}}_n \cdot S^{\mathrm{MHA}}_n,
    \label{eq:ffn_importance}
\end{equation}
\begin{equation}
    \mathcal{T}^{\mathrm{FFN}}_{\mathrm{active}}=\mathrm{TopK}(\{S^{\mathrm{FFN}}_n\}_{n=1}^N, M).
    \label{eq:ffn_selection}
\end{equation}

\subsection{Multi-Stage Delayed Token Pruning}
\label{sec:msdp}
For clarity, we denote $l$ as the layer index hereafter, and all variables in \cref{sec:pap,sec:ltp} are instantiated layer-wise, \emph{e.g.} $\mathcal{T}^l, \mathcal{T}^l_{\mathrm{active}}$ for the candidate and active sets, and $M^l$ for the number of active tokens.

As observed in \cref{fig:motivation}(a), tokens exhibit diminishing update magnitudes and increasing stability with depth, motivating a non-uniform token skipping strategy across layers. Moreover, prior studies suggest that the number of informative tokens do not scale with sequence length in long-context settings \cite{gemfilter, sliminfer}. Therefore, we adopt a fixed token budget per layer, rather than a fixed pruning ratio, enabling more consistent preservation of critical information. Concretely, the number of activate tokens is constrained by $M^l=\min (|\mathcal{T}^l|, M^l_{\mathrm{fixed}})$, where $M^l_{\mathrm{fixed}}$ is a predefined budget.

Furthermore, while keeping all context tokens as candidates intuitively preserves maximum information flow, doing so in deep layers incurs an increasingly imbalanced trade-off between probing overhead and computational savings, especially under fixed token budgets. 
More critically, due to the normalization property of attention, retaining excessive tokens in long sequences induces a smoothing effect on score distributions, undermining reliable top-token selection. As shown in \cref{fig:motivation}(c), 90\% of the attention mass can be captured by a nearly fixed number of tokens across deep layers, whereas achieving higher coverage demands rapidly more tokens as sequence length grows, revealing score dilution. In addition, \cref{fig:motivation}(d) shows that under non-uniform budgets, longer sequences exhibit lower top-token consistency between adjacent deep layers, indicating stronger interference from redundant tokens.

To address these issues, we introduce delayed pruning on top of token skipping to progressively remove redundant tokens. As illustrated in~\cref{fig:framework}(c), the skipping layers are divided into multiple stages, each followed by a pruning step at its last layer. Formally, let $\mathcal{L}^s=\{l_s,l_s\!+\!1,...,l_s\!+\!L_s\}$ denote the $L_s$ consecutive layers forming the $s$-th stage. All layers $l\in \mathcal{L}^s$ share the same candidate token set $\mathcal{T}^l$ of size $N_s$. At the end of stage $s$, \emph{i.e.} after layer $l_s\!+\!L_s$, tokens are pruned based on the last contribution scores $S_n^{l_s\!+\!L_s,\mathrm{MHA}}$ as \cref{eq:mha_importance}. This yields a reduced candidate set forwarding to subsequent layers:
\begin{equation}
    \mathcal{T}^l = \mathrm{TopK}(\{S_n^{l_s\!+\!L_s,\mathrm{MHA}}\}_{n=1}^{N_s}, N_{s+1}), ~\forall l\in \mathcal{L}^{s+1}.
    \label{eq:stage_prune}
\end{equation}

\section{Experimental Results and Analysis}

\subsection{Experimental Settings}

\textbf{Models.}
Following prior works \cite{ftp, gemfilter} and recent advances in long-context LLMs, we evaluate SPTS on three representative models: the large-scale LLaMA-3.1-8B-Instruct \cite{llama8b}, Qwen-2.5-7B-Instruct \cite{qwen7b}, and the compact openPangu-Embedded-1B-V1.1\cite{pangu}, denoted as LLaMA, Qwen and openPangu, respectively. We also extend SPTS to MLLMs, where long visual contexts dominate computation, using LLaVA-1.5-7B \cite{llavaimproved} and LLaVA-Next-7B \cite{llavanext}, referred to as LLaVA and LLaVA-Next, respectively.

\begin{table*}[!t]
\centering
\caption{Comparison of performance scores and TTFT acceleration ratios (TTFT ratio) by various methods for three representative LLMs on the LongBench dataset. The best results are highlighted in bold and the second-best ones are underlined.}
\vspace{-8pt}
\label{tab:longbench}
\setlength{\tabcolsep}{2pt}
{
\fontsize{8.6}{11}\selectfont
\begin{tabular}{lccccccccccccccccccc}
\toprule
\multicolumn{1}{c}{\multirow{4}{*}{Method}} & \multicolumn{2}{c}{\makecell{Single-Doc. \\QA}} & \multicolumn{3}{c}{\makecell{Multi-Doc.\\ QA}} & \multicolumn{4}{c}{Summarization} & \multicolumn{4}{c}{\makecell{Few-shot\\ Learning}} & \multicolumn{2}{c}{\makecell{Synthetic\\ Task}} & \multicolumn{2}{c}{\makecell{Code\\ Completion}} & \multicolumn{1}{c}{\multirow{4}{*}{\makecell{Avg. \\(\%)}}} & \multicolumn{1}{c}{\multirow{4}{*}{\makecell{TTFT\\ratio}}} \\
\cmidrule(lr){2-3}\cmidrule{4-6}\cmidrule(lr){7-10}\cmidrule(lr){11-14}\cmidrule{15-16}\cmidrule(lr){17-18}
 & \multicolumn{1}{c}{\rotatebox{90}{Qasper}} & \multicolumn{1}{c}{\rotatebox{90}{MQA}} & \multicolumn{1}{c}{\rotatebox{90}{HPQA}} & \multicolumn{1}{c}{\rotatebox{90}{2WiKi}} & \multicolumn{1}{c}{\rotatebox{90}{MuSiQue}} & \multicolumn{1}{c}{\rotatebox{90}{GovRep}} & \multicolumn{1}{c}{\rotatebox{90}{QMSum}} & \multicolumn{1}{c}{\rotatebox{90}{MNews}} & \multicolumn{1}{c}{\rotatebox{90}{VCSum}} & \multicolumn{1}{c}{\rotatebox{90}{TREC}} & \multicolumn{1}{c}{\rotatebox{90}{TQA}} & \multicolumn{1}{c}{\rotatebox{90}{SAMSum}} & \multicolumn{1}{c}{\rotatebox{90}{LSHT}} & \multicolumn{1}{c}{\rotatebox{90}{Count}} & \multicolumn{1}{c}{\rotatebox{90}{PassR}} & \multicolumn{1}{c}{\rotatebox{90}{LCC}} & \multicolumn{1}{c}{\rotatebox{90}{RepB-p}} &  \\ 
\midrule
 \multicolumn{20}{c}{\textit{LLaMA-3.1-8B-Instruct}}\\
\midrule
Full Model     & 45.16           & 54.71          & 55.64          & 44.73          & 30.66            & 34.99           & 25.26          & 27.17          & 17.16          & 72.50          & 91.66          & 43.98           & 46.50          & 7.00           & 99.50           & 63.02          & 56.01           & 47.98           & 1.00$\times$ \\
\midrule
LazyLLM     & \textbf{45.56}  & 53.17          & 53.47          & 45.50          & 30.44            & 32.76           & 24.84          & 26.19          & 16.99    & 71.00          & 91.25          & 43.28           & 46.00 & \textbf{7.50}  & \textbf{99.50}  & 58.66          & 55.17           & 47.13     & \underline{1.63$\times$}\\
MInference  & 44.59           & 52.64          & 53.51          & \textbf{47.30}    & 28.08            & \textbf{35.25}  & \textbf{25.51}    & \textbf{27.29} & \textbf{17.43} & \textbf{72.50} & 91.42    & 43.58     & \underline{46.50} & 6.55           & 95.50           & 63.30 & \underline{57.41}     & 47.55          & 0.47$\times$ \\
FlexPrefill & \underline{45.18}           & \textbf{55.52} & \textbf{56.95}    & 42.75          & \underline{32.14}      & \underline{34.55}           & \underline{25.16} & 26.86          & 17.39          & 71.00          & 90.74          & 43.50           & 42.00          & 3.16           & 81.50           & 62.93    & \textbf{60.89}  & 46.60           & 0.96$\times$ \\
SlimInfer   & \underline{45.18}     & 53.72    & 54.34 & 43.68 & 30.69   & 34.40     & 24.60          & \underline{27.28}    & 16.94          & 71.00    & \underline{91.64} & \textbf{44.63}  & 45.50          & 6.40     & 98.50     & \underline{63.38}          & 56.63           & \underline{47.56}   & 1.62$\times$\\
PyramidInfer & 44.02    & 54.22     &  53.75    & 44.49     & 30.39     & 33.98     & 24.64     & 26.06     & \underline{17.41}     & 68.00     & 90.98    & 43.58      & 46.00    & 5.97   & \textbf{99.50}     & \textbf{63.57}     & 57.04     & 47.27     & 1.19$\times$\\
FTP & 37.79    & 48.49     &  49.68    & 40.08     & 29.41     & 22.12     & 23.75     & 24.14     & 14.32     & 64.50     & 87.04    & 39.05      & 37.00    & 1.85   & 98.50     & 51.87     & 45.41     & 42.06     & 1.32$\times$\\
GemFilter & 44.65    & 52.48     &  \underline{55.84}    & 45.42     & \textbf{32.15}     & 33.53     & 22.12     & 27.11     & 16.56     & 70.50     & \textbf{92.05}    & 43.19      & \textbf{50.00}    & 5.31   & \textbf{99.50}     & 59.90     & 50.20     & 47.09     & 1.31$\times$ \\
\textbf{SPTS (Ours)} & 44.98    & \underline{55.37}     &  53.66    & \underline{46.69}     & 30.43     & 33.49     & 24.73     & 27.10     & 16.76     & \textbf{72.50}     & \underline{91.64}    & \underline{44.23}      & \underline{46.50}    & \underline{7.24}   & 99.00     & 63.14     & 55.20     & \textbf{47.80}     & \textbf{1.68$\times$}\\
\midrule
 \multicolumn{20}{c}{\textit{Qwen-2.5-7B-Instruct}}\\
\midrule
Full Model     & 43.92           & 52.76          & 57.97          & 46.56          & 30.16            & 31.78           & 23.36          & 24.30          & 16.05          & 72.50          & 88.64          & 45.64           & 43.00          & 8.00           & 100.0          & 60.44          & 66.84           & 47.76            & 1.00$\times$ \\
\midrule
LazyLLM     & 39.79           & 45.71          & 53.30          & 42.58          & 28.94            & 31.16           & 23.08          & 23.28          & 15.61          & 66.50          & 87.67          & 45.31           & 42.25    & 6.59           & \textbf{100.0} & 57.46          & 63.89           & 45.48          & 1.28$\times$  \\
MInference  & 43.61  & \textbf{52.32} & \textbf{57.59} & 45.31    & \underline{30.48}   & \underline{31.72}     & \underline{23.36}    & \textbf{24.10}          & 15.85          & \textbf{72.50}    & \textbf{89.61} & \textbf{46.33}     & 42.10          & \textbf{9.00}     & 91.50           & 59.88 & \textbf{66.84}  & 47.18     & 0.41$\times$ \\
FlexPrefill & 41.87           & 50.89          & 56.31          & 41.53          & 29.43      & \textbf{31.92}           & \textbf{23.66}    & 23.92    & \underline{15.92}    & 71.50          & 88.93          & 45.73  & 33.80          & 3.00           & 75.50           & \textbf{60.08}    & 62.47           & 44.50           & 0.90$\times$ \\
SlimInfer   & \underline{43.70}     & \underline{52.14}    & \underline{57.57}    & \underline{47.01} & 27.83            & 31.32  & 23.24 & 23.85 & \textbf{15.96} & \underline{72.00} & 88.56    & 45.55           & \textbf{43.50} & \textbf{9.00}  & 99.00     & 59.74          & \underline{66.36}     & \underline{47.43}  & \underline{1.34$\times$} \\
PyramidInfer & 37.09    & 45.85     &  54.88    & 44.88     & 29.01     & 30.23     & 23.05     & 22.82     & 15.34     & 63.50     & \underline{89.43}    & 46.26      & 42.25    & 8.50   & \textbf{100.0}     & \underline{59.99}     & 66.25     & 45.84     & 1.16$\times$\\
FTP & 41.09    & 48.37     &  53.65    & 36.45     & 20.75     & 28.05     & 22.23     & 21.99     & 15.33     & 65.50     & 89.23    & 43.88      & 36.50    & 6.50   & 98.50     & 59.57     & 64.69     & 44.25    & \textbf{1.36$\times$}\\
GemFilter & 42.28    & 44.34     &  52.12    & 46.54     & 22.82     & 30.46     & 19.79     & 23.85     & 15.22     & 68.00     & 88.31    & 43.26      & 41.25    & 5.50   & 97.50     & 59.79     & 65.05     & 45.06     & 1.30$\times$\\
\textbf{SPTS (Ours)} & \textbf{44.06}    & 51.29     &  56.66    & \textbf{47.37}     & \textbf{32.44}     & 30.92     & 22.96     & \underline{23.94}     & 15.72     & 70.50     & 89.04    & \underline{46.19}      & \underline{43.25}    & 8.00   & \textbf{100.0}     & 59.68     & 65.47     & \textbf{47.50}     & \textbf{1.36$\times$}\\
\midrule
 \multicolumn{20}{c}{\textit{openPangu-Embedded-1B-V1.1}}\\
\midrule
Full Model     & 27.85           & 44.15          & 32.17          & 34.75          & 17.78            & 26.62           & 19.68          & 22.78          & 14.56          & 67.00          & 79.70          & 38.96           & 33.17          & 2.62           & 15.50          & 25.37          & 25.80           & 31.09           & 1.00$\times$ \\
\midrule
PyramidInfer & \underline{25.05}    & 39.21     & \underline{30.05}     & \underline{30.87}     & \underline{15.54}     & 24.64     & 18.88     & 21.42     & 14.32     & \underline{64.50}     & \textbf{79.10}    & \textbf{38.59}      & \textbf{34.31}    & 1.33   & 14.50     & 23.78     & \underline{25.21}    & \underline{29.49}     & 0.83$\times$\\
FTP & 23.03    & \underline{39.22}     & 23.78     & 25.61     & 10.31     & 23.62     & \textbf{19.82}     & 19.99     & 14.33     & 60.00     & 76.03    & 36.00      & 28.57    & \underline{2.17}   & 14.00     & 20.28     & 18.09     & 26.76     & 0.87$\times$\\
GemFilter & 21.88    & 37.96     & 24.36     & 29.09     & 13.36     & \underline{24.66}     & 18.38     & \underline{22.67}     & \textbf{14.74}     & 62.00     & 77.81    & 35.16      & \underline{34.29}    & \textbf{2.64}   & \textbf{24.25}     & \textbf{25.83}     & \textbf{28.19}     & 29.25     & \underline{1.24$\times$}\\
\textbf{SPTS (Ours)} & \textbf{26.44}    & \textbf{43.83}     & \textbf{32.56}     & \textbf{36.22}     & \textbf{16.50}     & \textbf{24.82}     & \underline{19.02}     & \textbf{22.72}     & \underline{14.63}     & \textbf{66.50}     & \underline{79.05}    & \underline{38.32}      & 33.50    & 1.00   & \underline{15.00}     & \underline{25.49}     & 24.96     & \textbf{30.62}     &  \textbf{1.32$\times$}\\
\bottomrule
\end{tabular}
\vspace{-6pt}
}
\end{table*}

\textbf{Datasets and Evaluation Metrics.}
We conduct experiments on LongBench \cite{longbench}, a comprehensive benchmark containing 17 sub-datasets across 6 long-context task categories.
Following the official protocols, we report F1 for the Single-/Multi-Document QA tasks including Qasper, MultiFieldQA (MQA), HotpotQA (HPQA), 2WikiMultihopQA (2WiKi), MuSiQue; Rouge-L \cite{rougel} for Summarization tasks including  GovReport (GovRep), QMSum, MultiNews (MNews) and VCSUM; exact match Accuracy (\%) for Synthetic tasks including PassageCount (Count), PassageRetrieval (PassR); and Edit Sim (Levenshtein distance) \cite{edit_sim} for Code Completion tasks including LCC, RepoBench-P (RepB-p). For Few-shot Learning tasks, we present classification Accuracy (\%) on TREC and LSHT, F1 on TriviaQA (TQA), and Rouge-L on SAMSum.
We also report the average score over all subsets, denoted as Avg. (\%). To evaluate inference efficiency, following \cite{sliminfer}, we measure time to first token (TTFT) and end-to-end (E2E) latency for decoding 16 tokens. The corresponding Acceleration Ratios are computed as the latency of the full model divided by that of each method.
For MLLMs, following common practice \cite{pyramiddrop, sparsevlm}, we evaluate under fixed efficiency on 5 benchmarks: MME \cite{mme}, MMBench (MMB) \cite{mmbench}, GQA \cite{gqa}, TextVQA \cite{textvqa}, and the image subset of ScienceQA (SQA-I) \cite{sqa}. We report the summed perception and cognition score for MME, Accuracy (\%) for the remaining benchmarks, and their averaged score (Avg.).

\begin{figure*}[!t]
  \begin{center}
    \centerline{\includegraphics[width=0.95\textwidth]{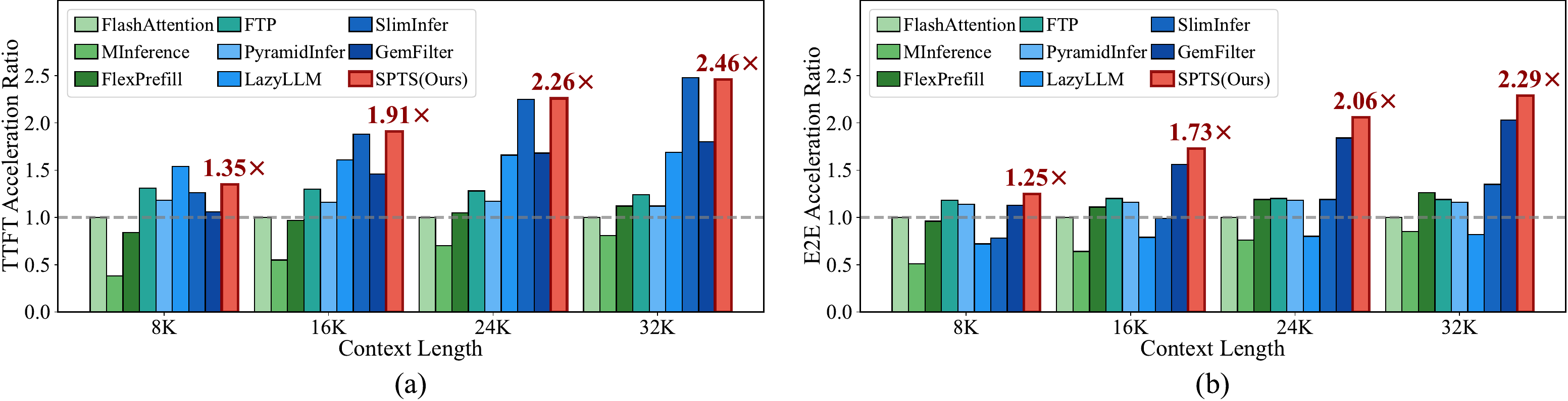}}
    \vspace{-10pt}
    \caption{
      Comparison of (a) TTFT acceleration ratios and (b) E2E acceleration ratios with different methods across distinct context lengths on LLaMA3.1-8B-Instruct.
    }
    \label{fig:ttft_and_e2e}
  \end{center}
  \vspace{-18pt}
\end{figure*}

\textbf{Implementation Details.}
We implement all methods with the Transformers library \cite{huggingface_transformers} and use the same CUDA/CANN runtime and operator kernels to ensure a fair comparison. All methods remain compatible with FlashAttention \cite{flashattention2}, which is adopted as the acceleration baseline. For FlexPrefill \cite{flexprefill} and MInference \cite{minference}, we retain their official Triton-based \cite{triton} implementations to measure practical speedups.
For token skipping and delayed pruning, we follow prior work \cite{sliminfer,pyramiddrop} in predefining the start layers for computation reduction. Then we evenly partition the layers into stages, and empirically reduce the token budget uniformly across stages to reach the target acceleration ratio, which ensures fair comparison of accuracy under comparable efficiency.
For long-context LLMs, we adopt a four-stage configuration. According to the total number of layers, for LLaMA (32 layers), token skipping is applied from layer 10 to 28; for Qwen (28 layers), from layer 9 to 24; and for openPangu (26 layers), from layer 11 to 24.
For MLLMs, the skipping range and stage number are adjusted to align with distinct computational budgets, with earlier skipping due to the higher redundancy and lower budgets of visual tokens \cite{pyramiddrop,fastv}.
Detailed per-layer budgets are provided in the \emph{Appendix}.
In the LTP module, we set the hyperparameters $(D_{\mathrm{low}}, r)$ to (512, 192), (1024, 256), (1024, 128), (256, 64) for LLaMA, Qwen, openPangu, and LLaVA/LLaVA-Next, respectively, according to their FFN scales.
For \cref{eq:channel_importance}, all LLMs adopt the 200 Qasper samples for calibration, with $\rho$ fixed to 0.2. For MLLMs, we sample 200 examples from the official Mix665K dataset \cite{llavaimproved}, and, given the relatively shorter visual contexts, we set $\rho$ to 0.5.
Experiments are conducted on one NVIDIA A800 GPU (80 GB) for LLaMA and Qwen, one Ascend 910B2 NPU (64GB) for openPangu, and one NVIDIA RTX A6000 GPU (48 GB) for LLaVA and LLaVA-Next. Additional ablations on hyperparameters and alternative configurations are provided in the \emph{Appendix}.

\subsection{Main Results on Accuracy-Efficiency Trade-off}

We compare our proposed SPTS with state-of-the-art long-context LLM acceleration methods. MInference \cite{minference} and FlexPrefill \cite{flexprefill} explore sparse attention, PyramidInfer \cite{pyramidinfer}, GemFilter \cite{gemfilter}, LazyLLM \cite{lazyllm} and SlimInfer \cite{sliminfer} are token pruning methods, and FTP \cite{ftp} implements token skipping. Since most methods primarily target speedup in prefilling, we conduct comparisons under comparable average TTFT acceleration ratios for fairness.

\begin{table}[!t]
\centering
\caption{Comparison of performance by various methods for LLaVA-1.5-7B on 5 benchmarks. The best results are highlighted in bold and the second-best ones are underlined.}
\vspace{-8pt}
\label{tab:main_llava}
\setlength{\tabcolsep}{2pt}
{
\fontsize{8.6}{11}\selectfont
\begin{tabular}{lcccccc}
\toprule
\multicolumn{1}{c}{Method} & MME & SQA-I & MMB & GQA & TextVQA & Avg. \\
\hline
\rowcolor{gray!20}
\multicolumn{7}{c}{\textit{Upper Bound: 576 Tokens} (\textbf{100\%})} \\
Full Model & 1860.8 & 69.5 & 64.6 & 61.9 & 58.3 & 64.1 \\
\hline
\rowcolor{gray!20}
\multicolumn{7}{c}{\textit{Retain 192 Tokens}  \textcolor{green!50!black}{($\downarrow$ \textbf{66.7\%})}} \\
FastV & 1612.0 & 67.3 & 61.2 & 52.7 & 52.5 & 58.3 \\
LLaVA-PruMerge & 1632.0 & 67.9 & 59.6 & 54.3 & 54.3 & 58.9 \\
SparseVLM & 1721.0 & \underline{69.1} & 62.5 & 57.6 & 56.1 & 61.4 \\
PDrop & \underline{1803.8} & 68.5 & \textbf{64.1} & \underline{59.2} & \underline{56.9} & \underline{62.6} \\
\textbf{SPTS (Ours)} & \textbf{1811.8} & \textbf{69.2} & \underline{63.8} & \textbf{59.8} & \textbf{57.0} & \textbf{62.9} \\
\hline
\rowcolor{gray!20}
\multicolumn{7}{c}{\textit{Retain 128 Tokens} \textcolor{green!50!black}{($\downarrow$ \textbf{77.8\%})}} \\
FastV & 1490.0 & 60.2 & 56.1 & 49.6 & 50.6 & 53.9 \\
LLaVA-PruMerge & 1554.0 & 67.1 & 58.1 & 53.3 & 54.3 & 57.7 \\
SparseVLM & \textbf{1696.0} & 67.1 & 60.0 & \underline{56.0} & 54.9 & 59.7 \\
PDrop & 1636.9 & \textbf{69.6} & \underline{61.2} & 55.2 & \underline{55.2} & \underline{59.9} \\
\textbf{SPTS (Ours)} & \underline{1660.7} & \underline{69.3} & \textbf{63.1} & \textbf{57.0} & \textbf{56.0} & \textbf{60.9} \\
\hline
\rowcolor{gray!20}
\multicolumn{7}{c}{\textit{Retain 64 Tokens}  \textcolor{green!50!black}{($\downarrow$ \textbf{88.9\%})}} \\
FastV & 1256.0 & 51.1 & 48.0 & 46.1 & 47.8 & 47.6 \\
LLaVA-PruMerge & \textbf{1549.0} & 68.1 & 55.3 & 51.9 & \textbf{54.0} & \underline{56.9} \\
SparseVLM & 1505.0 & 62.2 & \underline{56.2} & \textbf{52.7} & 51.8 & 55.3 \\
PDrop & 1255.6 & \underline{68.2} & 43.7 & 45.0 & 47.8 & 49.9 \\
\textbf{SPTS (Ours)} & \underline{1532.7} & \textbf{69.7} & \textbf{58.2} & \underline{52.4} & \underline{51.9} & \textbf{57.4} \\
\bottomrule
\end{tabular}
\vspace{-6pt}
}
\end{table}


\begin{table}[!t]
\centering
\caption{Comparison of performance by various methods for LLaVA-Next-7B on 5 benchmarks. The best results are highlighted in bold and the second-best ones are underlined.}
\vspace{-8pt}
\label{tab:main_llava_next}
\setlength{\tabcolsep}{2pt}
{
\fontsize{8.6}{11}\selectfont
\begin{tabular}{lcccccc}
\toprule
\multicolumn{1}{c}{Method} & MME & SQA-I & MMB & GQA & TextVQA & Avg. \\
\hline
\rowcolor{gray!20}
\multicolumn{7}{c}{\textit{Upper Bound: 2880 Tokens} (\textbf{100\%})} \\
Full Model & 1846.6 & 70.2 & 67.9 & 64.3 & 61.2 & 65.9 \\
\hline
\rowcolor{gray!20}
\multicolumn{7}{c}{\textit{Retain 320 Tokens}  \textcolor{green!50!black}{($\downarrow$ \textbf{88.9\%})}} \\
FastV & \textbf{1661.0} & 62.8 & \underline{61.6} & 55.9 & \underline{55.7} & 59.1 \\
LLaVA-PruMerge & 1534.0 & 66.4 & 61.3 & 53.6 & 50.6 & 57.3 \\
SparseVLM & 1533.0 & 66.1 & 60.6 & \underline{56.1} & \textbf{58.4} & \underline{59.2} \\
PDrop & 1470.3 & \textbf{68.8} & 56.6 & 51.8 & 49.8 & 55.9 \\
\textbf{SPTS (Ours)} & \underline{1659.3} & \underline{67.9} & \textbf{62.5} & \textbf{57.7} & 52.7 & \textbf{60.0} \\
\bottomrule
\end{tabular}
\vspace{-6pt}
}
\end{table}

\textbf{On Accuracy.}
As shown in \cref{tab:longbench}, SPTS consistently achieves higher average scores while delivering more substantial TTFT speedups. Specifically, it outperforms the second-best methods by 0.24\%, 0.07\% and 1.13\% on LLaMA, Qwen and openPangu, respectively. Notably, while SlimInfer is competitive in both accuracy and TTFT acceleration, it suffers from pronounced degradation in decoding speed, as discussed below. MInference and FlexPrefill require online sparsity pattern extraction, incurring non-negligible computational overhead that is only amortized in ultra-long contexts \cite{minference}. As a result, despite competitive accuracy on certain datasets, the negative speedups (< 1$\times$) limit their applicability in general settings.
GemFilter and PyramidInfer prune tokens directly during prefilling, potentially discarding critical contextual information and causing performance degradation. FTP supports token skipping but relies on outdated, redundancy-affected selection criteria, thus leading to significant accuracy loss under high acceleration ratios.
Overall, by enabling more comprehensive token skipping and leveraging more informative token selection signals, SPTS preserves and extracts critical information from long sequences more effectively without heavy auxiliary operations. The consistent improvements across models of varying structures, scales, and hardware platforms further highlight the strong generalization capability of our method.

\textbf{On Efficiency.}
Following \cite{sliminfer}, we sample input sequences of four different lengths ($8$K, $16$K, $24$K and $32$K) from LongBench \cite{longbench} to evaluate inference efficiency. For each context length, 20 distinct sequences are sampled, and the corresponding latencies are recorded and averaged following a warmup period.
We evaluate both TTFT and E2E acceleration ratios across different methods, with the full model using dense FlashAttention \cite{flashattention2} as the baseline.
As shown in \cref{fig:ttft_and_e2e}, SPTS consistently accelerates both the prefilling and decoding phases, with speedups increasing as the context length grows. At a length of $32$K, it achieves 2.46$\times$ and 2.29$\times$ acceleration for TTFT and E2E latencies, respectively. Importantly, these efficiency gains are obtained while maintaining state-of-the-art average accuracy in \cref{tab:longbench}, highlighting its significant efficiency potential in long-context inference scenarios.

\begin{table}[t]
\centering
\caption{Ablation of the PAP, LTP and MSDP modules on TTFT acceleration ratios across context lengths with LLaMA.}
\vspace{-6pt}
{
\setlength{\tabcolsep}{6pt}
\fontsize{8.6}{11}\selectfont
\newcolumntype{L}[1]{>{\raggedright\arraybackslash}p{#1}}
\begin{tabular}{L{2.9cm}|cccc}
\toprule
\multicolumn{1}{c|}{Method} & $8$K & $16$K & $24$K & $32$K \\
\midrule
Full Model & 1.00$\times$ & 1.00$\times$ & 1.00$\times$ & 1.00$\times$ \\
+ PAP & 1.08$\times$ & 1.24$\times$ & 1.35$\times$ & 1.44$\times$ \\
+ PAP + LTP & 1.26$\times$ & 1.70$\times$ & 1.98$\times$ & 2.15$\times$ \\
+ PAP + LTP + MSDP & \textbf{1.35$\times$} & \textbf{1.91$\times$} & \textbf{2.26$\times$} & \textbf{2.46$\times$} \\
\bottomrule
\end{tabular}
\vspace{-6pt}
}
\label{tab:ablation_main_speed}
\end{table}

\begin{table}[t]
\centering
\caption{Effect of the PAP module on KV Cache memory (GB) savings across varying context lengths with LLaMA.}
\vspace{-6pt}
{
\setlength{\tabcolsep}{6pt}
\fontsize{8.6}{11}\selectfont
\newcolumntype{L}[1]{>{\raggedright\arraybackslash}p{#1}}
\begin{tabular}{L{3.5cm}|cccc}
\toprule
\multicolumn{1}{c|}{Method} & $8$K & $16$K & $24$K & $32$K \\
\midrule
Ours w/o PAP (Full Cache)         & 4.00 & 8.00 & 12.00 & 16.00 \\
Ours with PAP        & 2.81 & 4.13 & 5.38  & 6.63 \\
Memory Saving (\%) & \textbf{29.7} & \textbf{48.4} & \textbf{55.2} & \textbf{58.6} \\
\bottomrule
\end{tabular}
\vspace{-6pt}
}
\label{tab:ablation_pap_memory}
\end{table}

\textbf{On Generalization to MLLMs.}
Our SPTS is designed as a modality-general framework, with its motivation and observations holding across modalities. Since recent MLLMs adopt the same long-sequence processing paradigm as LLMs, we further evaluate SPTS on the LLaVA model family \cite{llavaimproved, llavanext} to validate its effectiveness for visual understanding, treating only the visual tokens as candidates for computation reduction. We compare against state-of-the-art visual token pruning methods, including FastV \cite{fastv}, LLaVA-PruMerge \cite{llavaprumerge}, SparseVLM \cite{sparsevlm} and PDrop \cite{pyramiddrop}. Following common practice, methods are evaluated under the same average number of active tokens across all layers, which ensures matched efficiency.
As shown in~\cref{tab:main_llava}, on LLaVA, which natively uses a fixed length of 576 visual tokens, SPTS achieves average performance gains of 0.3\%, 1.0\% and 0.5\% when retaining 192, 128 and 64 tokens, respectively. 
We further evaluate on LLaVA-Next, which supports higher image resolution with up to 2{,}880 visual tokens. As reported in~\cref{tab:main_llava_next}, under an 88.9\% token reduction, SPTS still yields an average improvement of 0.8\%. This stems from performing token skipping prior to pruning, which effectively preserves visual contextual information in the early layers of the network~\cite{vdc,pade}. Notably, these gains are achieved with no additional visual-specific optimization, further highlighting the generalization capability of our method.

\subsection{Ablation Study}
\label{sec:exp_ablation}

\textbf{On Main Components.}
We evaluate the efficiency contributions of the main components: the Partial Attention Probing (PAP) for MHA, the Low-rank Transformation Probing (LTP) for FFN, and the Multi-Stage Delayed Pruning (MSDP) strategy. As shown in \cref{tab:ablation_main_speed}, sequentially incorporating PAP and LTP increases the amount of skipped computation, resulting in cumulative speedups of 1.44$\times$ and 2.15$\times$, respectively. The introduction of MSDP further improves efficiency by reducing the overhead of token selection, ultimately yielding a total speedup of 2.46$\times$ for a $32$K context length.

\textbf{On PAP.}
We analyze the effect of the PAP design. As shown in \cref{tab:ablation_pap_memory}, incorporating PAP reduces the KV Cache memory footprint while maintaining token forward propagation in prefilling. For a $32$K context, it achieves up to 58.6\% memory savings. This demonstrates that our method not only exploits FFN-level token skipping as FTP \cite{ftp} but also provides an advantage in memory efficiency.

\begin{table}[t]
\centering
\caption{Effect of low-rank proxy network on LLaMA. Results are reported on five representative LongBench subsets and the average score (Avg.) on the full dataset.}
\vspace{-6pt}
{
\fontsize{8.6}{11}\selectfont
\setlength{\tabcolsep}{3pt}
\newcolumntype{L}[1]{>{\raggedright\arraybackslash}p{#1}}
\begin{tabular}{L{1.7cm}|c c c c c | c}
\toprule
\multicolumn{1}{c|}{\multirow{2}{*}{\makecell{Skipping\\Criterion}}} & \multicolumn{5}{c|}{Dataset} & \multirow{2}{*}{\makecell{Avg.\\(\%)}}\\
\cline{2-6}
                                             & MQA            & HPQA           & SAMSum         & Count         & RepB-p         & \\
\midrule
$S_n^{\mathrm{MHA}}$                         & 52.96          & 53.39          & 44.23          &  5.59         & 52.79          & 46.31 \\
$C_n^{\mathrm{FFN}}\cdot S_n^{\mathrm{MHA}}$ & \textbf{53.71} & \textbf{54.11} & \textbf{44.64} & \textbf{7.47} & \textbf{53.26} & \textbf{46.85} \\
\bottomrule
\end{tabular}
\vspace{-4pt}
}
\label{tab:ablation_ffn_metric}
\end{table}

\begin{table}[t]
\centering
\caption{Effect of Delayed Pruning (DP) on LLaMA across five LongBench subsets and average (Avg.) on the full dataset.}
\vspace{-6pt}
{
\fontsize{8.6}{11}\selectfont
\setlength{\tabcolsep}{3pt}
\newcolumntype{L}[1]{>{\raggedright\arraybackslash}p{#1}}
\begin{tabular}{L{1.7cm}|c c c c c | c}
\toprule
\multicolumn{1}{c|}{\multirow{2}{*}{Method}} & \multicolumn{5}{c|}{Dataset} & \multirow{2}{*}{\makecell{Avg.\\(\%)}}\\
\cline{2-6}
                & MQA            & HPQA           & SAMSum         & Count         & RepB-p         & \\
\midrule
Ours w/o DP     & 54.99          & 52.51          & 43.72          &  4.32         & 52.39          & 46.11 \\
Ours with DP      & \textbf{55.37} & \textbf{53.66} & \textbf{44.23} & \textbf{7.24} & \textbf{55.20} & \textbf{47.80} \\
\bottomrule
\end{tabular}
\vspace{-2pt}
}
\label{tab:ablation_pruning}
\end{table}

\textbf{On LTP.}
We further investigate the role of the proxy network in the LTP module. To isolate its effect, we only apply token skipping to FFN blocks, with 40\% of tokens skipped in all deep layers. As shown in \cref{tab:ablation_ffn_metric}, relying solely on the attention-based contribution scores from the preceding MHA block, \emph{i.e.} \cref{eq:mha_importance}, provides a reasonable basis for token selection. However, incorporating the low-rank proxy network $f(\cdot)$ with a conditioned transformation criterion as \cref{eq:ffn_importance} introduces an explicit prediction of token-wise response to the target FFN block, which effectively preserves tokens that are expected to undergo substantial updates. This provides a more accurate and robust selection for active tokens, and finally results in consistent performance gains across tasks, with an average score improvement of 0.54\%.

\textbf{On MSDP.} We assess the effect of delayed pruning, as shown in \cref{tab:ablation_pruning}. By applying pruning on top of token skipping, redundant tokens are gradually removed, improving both the accuracy and stability of token selection. This is particularly beneficial for long-sequence tasks such as HPQA and Count, where pruning mitigates the smoothing of selection scores induced by extended contexts. Overall, it yields an average score improvement of 1.69\%.

\section{Conclusion}
In this work, we investigate token skipping-based acceleration for long-context LLM inference, and propose a novel training-free framework, dubbed SPTS. By introducing the Partial Attention Probing (PAP) and Low-rank Transformation Probing (LTP) modules, SPTS achieves comprehensive acceleration of the model’s predominant computational components while leveraging a self-predictive mechanism for more accurate token skipping. In addition, a Multi-Stage Delayed Pruning (MSDP) strategy is incorporated to allocate token budgets effectively and mitigate interference from redundant tokens. 
Extensive experiments on three representative long-context LLMs and two hardware platforms demonstrate that SPTS consistently improves inference latency while maintaining model accuracy, \emph{e.g.} achieving 2.46$\times$ TTFT and 2.29$\times$ E2E speedups, highlighting its effectiveness and generalizability in long-context scenarios.
Evaluation on MLLMs further validate its generalization capability for long visual contexts, establishing a promising baseline for future modality-specific optimizations.

\newpage
\clearpage

\section*{Acknowledgments} 
This work was supported in part by the Beijing Natural Science Foundation under Grant 4242044 and  L259044, the CCF Baidu Open Fund, the Research Program of the State Key Laboratory of Virtual Reality Technology and Systems, the Fundamental Research Funds for the Central Universities and the Academic Excellence Foundation of BUAA for PhD Students.

\bibliographystyle{ACM-Reference-Format}
\bibliography{main}

\clearpage
\setcounter{page}{1}
\setcounter{section}{0}
\renewcommand\thesection{\Alph{section}}
\setcounter{figure}{0}
\setcounter{table}{0}
\setcounter{equation}{0}
\renewcommand{\thefigure}{\Alph{figure}}
\renewcommand{\thetable}{\Alph{table}}
\renewcommand{\theequation}{\alph{equation}}

\twocolumn[
\begin{center}
{\huge\bfseries
Appendix of\\[4pt]
Probe and Skip: Self-Predictive Token Skipping \\
for Efficient Long-Context LLM Inference
}
\end{center}
\vspace{20pt}
]

In this document, we additionally provide a detailed description of inference procedure with SPTS in~\cref{sec:pseudocode}, more implementation details in~\cref{sec:more_implementation_details}, further discussions on the advantage of SPTS in terms of KV Cache flexibility in~\cref{sec:further_on_cache}, as well as more comprehensive experimental results and analysis in~\cref{sec:more_ablations}.
Specifically, we conduct further ablation studies on the scale of the low-rank proxy network in~\cref{sec:ablation_ffn_scale}, the query token length for partial attention probing in~\cref{sec:ablation_query_token_num}, the fusion design for conditional transformation in~\cref{sec:ablation_fusion_type}, the utilization of top-$\rho$ saliency values in~\cref{sec:ablation_top_rho}, the stage partitioning configurations in~\cref{sec:ablation_stage_partition} and the source and size of calibration data in~\cref{sec:ablation_calibration_data,sec:ablation_calibration_size}.

\section{Overall Algorithm of SPTS Inference Process}
\label{sec:pseudocode}
In~\cref{algo:inference}, we present the detailed procedure of LLM inference with SPTS for each auto-regressive forward propagation. Lines $8\sim 18$ describe the MHA token skipping with Partial Attention Probing (PAP), as illustrated in Fig.~2(a). Lines $24\sim 30$ present the FFN token skipping with Low-rank Transformation Probing (LTP), as displayed in Fig.~2(b). Lines $35\sim 37$ correspond to the delayed pruning strategy at each stage boundary, as depicted Fig.~2(c). For clarity, standard operations that are orthogonal to our method, such as layer normalization and positional embedding, are omitted.

\section{More Implementation Details}
\label{sec:more_implementation_details}
In Sec.~4.1, we describe the design principles for stage division and token budget reduction across stages. Specifically, we follow the compared methods \cite{sliminfer,pyramiddrop} to predefine the starting layers for computation reduction, and keep the last few layers intact without token skipping due to their observed instability as in Fig.~3(a). In most cases, we uniformly divide the skipping layers into four stages, where the active token budget is progressively reduced in a linear-like manner across stages, followed by an additional one-step pruning at each stage ending layer. 
Here, we detail the resulting layer-wise configurations for each model in~\cref{tab:layer_config}. Taking LLaMA (32 layers) as an example, token skipping is applied from layer 10 to 28, which is divided into four stages, each ending at layers $(13, 18, 23, 28)$. The fixed active token budgets $M_\mathrm{fixed}^l$ for each stage are set to (9K, 7K, 4K, 2K) (1K = 1024 tokens), and the candidate token set is further pruned by an additional 1K tokens beyond the active ones at the end of each stage.
For the LLaVA model family, skipping layers is scheduled earlier to align with the rather aggressive efficiency targets adopted by compared methods. Especially, under the stringent constraint of reducing an average of 88.9\% tokens per layer, \emph{i.e.} LLaVA-1.5-7B (64 Tokens) and LLaVA-Next-7B (320 Tokens) in~\cref{tab:layer_config}, we complete token skipping and pruning within the first 16 layers, and accordingly partition the layers into only three stages. 
Notably, such adjustments are made solely to ensure a fair comparison of accuracy at comparable efficiency levels. As prior works \cite{sliminfer, pyramiddrop} evaluate models under varying efficiency levels, different models are accordingly assigned distinct configurations. 

\begin{figure}[t]
  \begin{center}
    \centerline{\includegraphics[width=\linewidth]{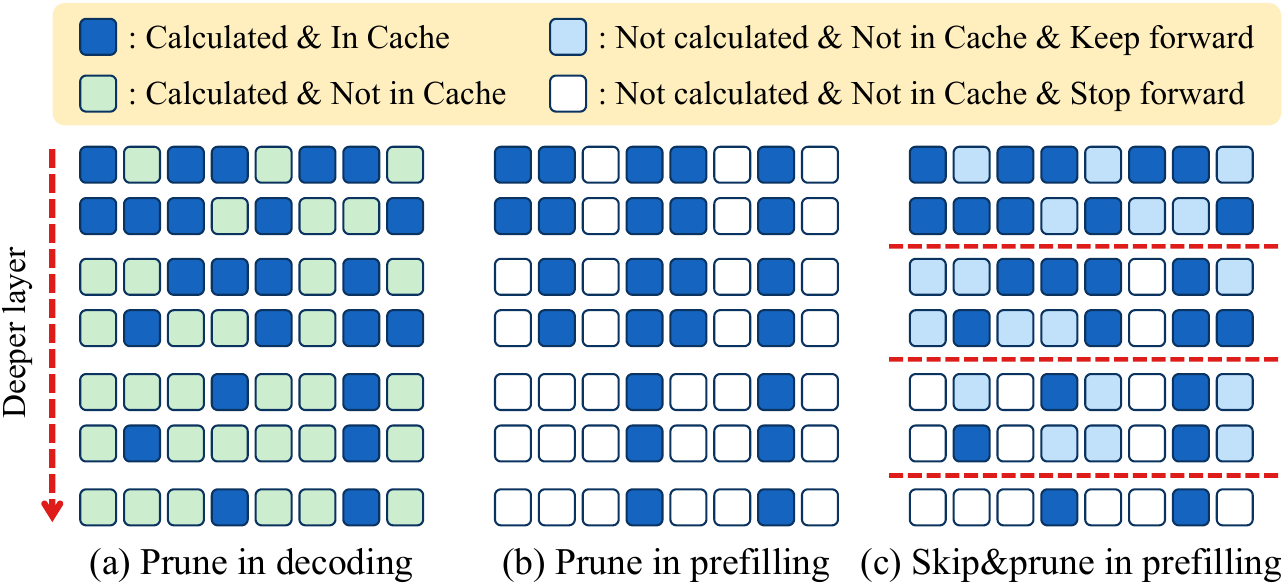}}
    \caption{Illustration of different schemes in KV Cache management. (a) Decoding-phase pruning computes all tokens, offering no speedup in prefilling. (b) Prefilling-phase pruning discard tokens to reduce computation, but cached tokens exist only in the early consecutive layers, leading to noticeable information loss. (c) SPTS combines skipping with delayed pruning, allowing skipped tokens to be recomputed and cached later. It achieves the same computation costs as (b) while preserving richer contextual information.}
    \label{fig:illu_cache}
  \end{center}
\end{figure}

\begin{table}[t]
\centering
\caption{Effect of $D_{\mathrm{low}}$ and $r$ on per-token per-projection FLOPs and accuracy for LLaMA across five representative LongBench subsets. `/' denotes no decomposition on the corresponding dimension. The default setting is underlined.}
{
\fontsize{8.6}{11}\selectfont
\setlength{\tabcolsep}{5pt}
\begin{tabular}{c c|r|cccccc}
\toprule
$D_{\mathrm{low}}$ & $r$ & \makecell[c]{FLOPs}         & Qasper & HPQA & TQA  & LSHT & LCC \\
\midrule
512 & 128  & 590K                        & 43.67 & 53.60 & 91.17 & 45.00 & 57.49 \\
512 & 256  & 1180K                       & 44.00 & 54.36 & 92.04 & 45.50 & 57.76 \\
512 & / & 2097K                          & 43.85 & 55.18 & 92.03 & 46.00 & 58.36 \\
\underline{512} & \underline{192}& 855K  & 43.85 & 54.11 & 91.59 & 45.00 & 57.65 \\
256 & 192 & 836K                         & 43.17 & 53.38 & 91.33 & 45.50 & 57.55 \\
1536 & 192 & 1081K                       & 43.87 & 53.93 & 91.64 & 46.00 & 57.79 \\
/ & 192 & 2900K                          & 42.70 & 55.75 & 91.08 & 46.50 & 58.97 \\
\bottomrule
\end{tabular}
}
\label{tab:ablation_proxy_rank}
\end{table}

\begin{algorithm*}[t]
    \small
    \caption{\small Self-Predictive Token Skipping (SPTS)}
    \label{algo:inference}
    \setcounter{AlgoLine}{0}
    \linespread{1.2}\selectfont
    \LinesNumbered

    \KwIn{An LLM with $L$ layers, consisting of standard MHA blocks $\{\mathcal{F}^l_{\mathrm{MHA}}\}_{l=1}^L$ and FFN blocks $\{\mathcal{F}^l_{\mathrm{FFN}}\}_{l=1}^L$; pre-constructed low-rank proxy networks $\{f^l(\cdot)\}_{l=1}^L$; candidate token budgets $\{N_s\}$ and active token budgets $\{M^s_{\mathrm{fixed}}\}$ for multiple stages; input hidden states $\bm{X}$, KV Cache $Cache$.}

    \KwOut{Final hidden states $\bm{X}$ for next token prediction, updated KV Cache $Cache$.}

    \BlankLine
    
    \For{$l \gets 1$ \KwTo $L$}{
        \textcolor{gray}{\fontfamily{pcr}\selectfont {\# Get current sequence length. All tokens are initially candidate for skipping.}}
        
        $N\gets \mathrm{Get\_sequence\_length}(\bm{X})$ 
    
        $skip \gets \mathrm{Is\_prefill\_phase}()~\wedge~\mathrm{In\_skip\_layers}(l)$ 
        \textcolor{gray}{\fontfamily{pcr}\selectfont {\# Whether token skipping is enabled at layer $l$.}}

        $s\gets\mathrm{Get\_stage\_index}(l)$
        \textcolor{gray}{\fontfamily{pcr}\selectfont {\# Get the stage index associated with layer $l$.}}

        \BlankLine

        \textcolor{gray}{\fontfamily{pcr}\selectfont {\# Partial Attention Probing (PAP) for MHA.}} \\
        \If{$skip$}{
            \textcolor{gray}{\fontfamily{pcr}\selectfont {\# Compute key states for all tokens and query state for the last token.}}
        
            $\bm{K} \gets \bm{X}\bm{W}_K$,\quad$\bm{q} \gets \bm{X}_N\bm{W}_Q$

            $S^{\mathrm{MHA}} \gets \mathrm{Mean\_softmax\_attention}(\bm{q}, \bm{K})$
            \textcolor{gray}{\fontfamily{pcr}\selectfont {\# Predict token contribution scores with Eq.~(5).}}

            \textcolor{gray}{\fontfamily{pcr}\selectfont {\# Get token budget and select active token indices.}}
            
            $m\gets\min(N, M_{\mathrm{fix}}^s)$,\quad$\mathcal{T}_{\mathrm{active}}^{\mathrm{MHA}} \gets \mathrm{TopK}(S^{\mathrm{MHA}}, m)$
            
            $\hat{\bm{K}}\gets\bm{K}[\mathcal{T}_{\mathrm{active}},~:]$
            \textcolor{gray}{\fontfamily{pcr}\selectfont {\# Gather computed key states corresponding to active tokens.}}
            
            \textcolor{gray}{\fontfamily{pcr}\selectfont {\# Extract active tokens and compute reduced projections for query and value states.}}
            
            $\hat{\bm{X}}\gets\bm{X}[\mathcal{T}_{\mathrm{active}},~:]$,\quad$\hat{\bm{Q}}\gets\hat{\bm{X}}\bm{W}_Q$,\quad$\hat{\bm{V}}\gets\hat{\bm{X}}\bm{W}_V$

            $Cache \gets Cache\cup (\hat{\bm{K}}, \hat{\bm{V}})$
            \textcolor{gray}{\fontfamily{pcr}\selectfont {\# Save the active key-value pairs into KV Cache.}}

            $\hat{\bm{O}}\gets\mathcal{F}^{l}_{\mathrm{MHA}}(\hat{\bm{Q}}, \hat{\bm{K}}, \hat{\bm{V}})$
            \textcolor{gray}{\fontfamily{pcr}\selectfont {\# Perform reduced MHA computation over active tokens.}}

            $\bm{X}[\mathcal{T}_{\mathrm{active}}^{\mathrm{MHA}},~:]\gets \hat{\bm{X}} + \hat{\bm{O}}$
            \textcolor{gray}{\fontfamily{pcr}\selectfont {\# Updated active token representations via residual connection.}}
        }
        \Else{
            \textcolor{gray}{\fontfamily{pcr}\selectfont {\# Standard MHA computation with residual connection. $Cache$ is also updated.}}\\
            $\bm{X}\gets \bm{X} + \mathcal{F}^{l}_{\mathrm{MHA}}(\bm{X}, Cache)$
        }

        \BlankLine
        
        \textcolor{gray}{\fontfamily{pcr}\selectfont {\# Low-rank Transformation Probing (LTP) for FFN.}} \\
        \If{$skip$}{
            $\bm{o} \gets f^l(\bm{X})$,\quad$C^{\mathrm{FFN}}\gets \|\bm{o}\|_2$
            \textcolor{gray}{\fontfamily{pcr}\selectfont {\# Predict token-wise transformation with Eq.~(13).}}

            $S^{\mathrm{FFN}}\gets S^{\mathrm{MHA}} \times C^{\mathrm{FFN}}$
            \textcolor{gray}{\fontfamily{pcr}\selectfont {\# Compute the conditioned transformation scores with Eq.~(14).}}

            \textcolor{gray}{\fontfamily{pcr}\selectfont {\# Get token budget and select active token indices.}}
            
            $m\gets\min(N, M_{\mathrm{fix}}^s)$,\quad$\mathcal{T}_{\mathrm{active}}^{\mathrm{FFN}} \gets \mathrm{TopK}(S^{\mathrm{FFN}}, m)$

            \textcolor{gray}{\fontfamily{pcr}\selectfont {\# Select active tokens and perform reduced FFN computation.}}
            
            $\hat{\bm{X}}\gets \bm{X}[\mathcal{T}_{\mathrm{active}}^{\mathrm{FFN}},~:]$,\quad$\hat{\bm{O}}\gets \mathcal{F}^l_{\mathrm{FFN}}(\hat{\bm{X}})$

            $\bm{X}[\mathcal{T}_{\mathrm{active}}^{\mathrm{FFN}},~:]\gets \hat{\bm{X}} + \hat{\bm{O}}$
            \textcolor{gray}{\fontfamily{pcr}\selectfont {\# Updated active token representations via residual connection.}}
        }
        \Else{
            $\bm{X}\gets \bm{X} + \mathcal{F}_{\mathrm{FFN}}^l(\bm{X})$
            \textcolor{gray}{\fontfamily{pcr}\selectfont {\# Standard FFN computation with residual connection.}}
        }

        \BlankLine
        
        \textcolor{gray}{\fontfamily{pcr}\selectfont {\# Execute delayed pruning at stage boundaries.}}

        \If{$\mathrm{Is\_stage\_end}(l, s)$}{
            \textcolor{gray}{\fontfamily{pcr}\selectfont {\# Select candidate token indices for the next stage.}}
            
            $m\gets\min(N, N_{s+1})$,\quad$\mathcal{T}^{l+1}\gets\mathrm{TopK}(S^{\mathrm{MHA}}, m)$

            $\bm{X}\gets\bm{X}[\mathcal{T}^{l+1},~:]$
            \textcolor{gray}{\fontfamily{pcr}\selectfont {\# Discard the redundant tokens and shrink the hidden state sequence.}}
        }
    }

    \BlankLine

    \Return $\bm{X}$, $Cache$
    
\end{algorithm*}

\section{Further Remark on KV Cache Flexibility}
\label{sec:further_on_cache}
We summarize existing methods from the perspective of KV Cache management. As illustrated in \cref{fig:illu_cache}, SPTS exhibits an additional advantage by striking a favorable accuracy-efficiency trade-off through flexible KV Cache compression. Specifically, as described in Sec.~3.2, by incorporating PAP for token skipping in MHA, the key and value projections are computed only for active tokens, which are subsequently stored in the KV Cache. This naturally enables an online KV Cache compression during the prefilling phase. Meanwhile, benefiting from the residual connections, tokens skipped at current layer are not permanently discarded, yet remain eligible to participate in subsequent layers, where they can be recomputed and cached when necessary. 
Consequently, SPTS improves dual-phase efficiency while enabling more flexible KV cache management, ensuring that informative tokens can be adaptively preserved across layers and thereby delivering improved accuracy.

\begin{table*}[!t]
\centering
\caption{Token skipping and pruning configurations for various models. `K' denotes 1024 tokens.}
\label{tab:layer_config}
\setlength{\tabcolsep}{5pt}
{
\fontsize{8.6}{11}\selectfont
\newcolumntype{R}[1]{>{\raggedleft\arraybackslash}p{#1}}
\newcolumntype{L}[1]{>{\raggedright\arraybackslash}p{#1}}
\begin{tabular}{lcR{0.2cm}R{0.887cm}L{0.2cm}ccc}
\toprule
\makecell[c]{Model} & Total Layers & \multicolumn{3}{c}{Skipping Layers} & Stage Ending Layers & Active Token Budgets & Pruned Tokens \\
\midrule
LLaMA-3.1-8B-Instruct & 32 & & 10 $\sim$ 28 &  & (13, 18, 23, 28) & (9K, 7K, 4K, 2K) & 1K \\
Qwen-2.5-7B-Instruct  & 28 & & 9 $\sim$ 24 &  & (12, 16, 24, 20) & (13K, 10K, 7K, 4K) & 2K \\
openPangu-Embedded-1B-V1.1 & 26 & & 11 $\sim$ 22& & (13, 16, 19, 22) & (13K, 10K, 7K, 4K) & 2K \\
LLaVA-1.5-7B (192 Tokens) & 32 & & 2 $\sim$ 24 & & (6, 12, 18, 24) & (316, 243, 169, 96) & 32 \\
LLaVA-1.5-7B (128 Tokens) & 32 & & 2 $\sim$ 24 & & (6, 12, 18, 24) & (184, 144, 104, 64) & 32 \\
LLaVA-1.5-7B (64 Tokens) & 32 & & 2 $\sim$ 16 & & (6, 11, 16) & (88, 56, 24) & 16 \\
LLaVA-Next-7B (320 Tokens) & 32 & & 2 $\sim$ 16 & & (6, 11, 16) & (432, 272, 112) & 80 \\
\bottomrule
\end{tabular}
}
\end{table*}

\begin{table*}[t]
\centering
\caption{Effect of query token length for PAP on LLaMA. Results are reported on representative LongBench subsets and the average score (Avg.) on the full dataset. The default setting is underlined.}
{
\fontsize{8.6}{11}\selectfont
\setlength{\tabcolsep}{5pt}
\begin{tabular}{>{\centering}p{2.3cm}|ccccccccc|c}
\toprule
\multicolumn{1}{c|}{Query Length}       & MQA   & HPQA  & 2WiKi & MuSiQue & QMSum & TREC & LSHT & Count & LCC & Avg. (\%)\\
\midrule
64 & 54.33 & 52.67 & 43.95 & 31.31 & 24.29 & 63.00 & 45.00 & 7.38 & 63.32 & 47.00 \\
4  & 54.53 & 54.83 & 45.39 & 30.58 & 25.23 & 70.50 & 46.50 & 4.77 & 62.85 & 47.53 \\
\underline{1}  & 55.37 & 53.66 & 46.69 & 30.43 & 24.73 & 72.50 & 46.50 & 7.24 & 63.14 & \textbf{47.80} \\
\bottomrule
\end{tabular}
}
\label{tab:ablation_window}
\end{table*}

\begin{table*}[t]
\centering
\caption{Effect of various fusion strategies for the conditional transformation score on LLaMA. Results are reported on representative LongBench subsets and the overall average (Avg.) across the full dataset. The default setting is underlined.}
{
\fontsize{8.6}{11}\selectfont
\setlength{\tabcolsep}{5pt}
\begin{tabular}{>{\centering}p{2.3cm}|ccccccccc|c}
\toprule
\multicolumn{1}{c|}{Method}         & MQA   & HPQA  & 2WiKi & MuSiQue & QMSum & TREC & LSHT & Count & LCC & Avg. (\%)\\
\midrule
\makecell[l]{MaxNorm-Add}           & 55.27 & 53.82 & 45.46 & 29.13 & 24.60 & 72.00 & 45.50 & 6.84 & 62.45 & 47.48 \\
\makecell[l]{Std-Add}               & 55.00 & 53.63 & 44.72 & 28.99 & 24.63 & 71.50 & 45.50 & 6.88 & 62.57 & 47.40 \\
\makecell[l]{Sequential Select}     & 55.15 & 53.37 & 45.52 & 28.37 & 24.34 & 72.50 & 45.50 & 5.47 & 62.88 & 47.43  \\
\makecell[l]{\underline{Multiply}}  & 55.37 & 53.66 & 46.69 & 30.43 & 24.73 & 72.50 & 46.50 & 7.24 & 63.14 & \textbf{47.80} \\
\bottomrule
\end{tabular}
}
\label{tab:ablation_fusion_method}
\end{table*}

\section{More Experimental Results and Analysis}
\label{sec:more_ablations}

\subsection{Ablation on Scale of the Low-rank Proxy Network}
\label{sec:ablation_ffn_scale}
We investigate the effects of $D_{\mathrm{low}}$ and $r$ in Sec.~3.3, which jointly determine the scale of the proxy network used for token selection. Following the previous setup in Sec.~4.3, token skipping is applied only to FFN blocks. As shown in \cref{tab:ablation_proxy_rank}, increasing $D_{\mathrm{low}}$ and $r$ enhances the fidelity of the proxy network in approximating the original FFN, which enables more accurate prediction of token-wise transformations. This further leads to more precise selection of critical tokens, thus alleviating the performance degradation caused by reduced feed-forward computation. 
Conversely, larger values incur higher computational overhead for token selection, which we quantify in terms of FLOPs (Floating-Point operations) per token per projection in~\cref{tab:ablation_proxy_rank}. This establishes a trade-off between efficiency and fidelity in constructing the proxy network.

We also include settings that separately remove the $D_{\mathrm{low}}$ reduction and the $r$-based SVD decomposition, denoted as $(D_{\mathrm{low}}, r)=(/, 192)$ and $(512, /)$, respectively, revealing that preserving the rank $r$ is more critical to the effectiveness of the proxy network. 
In particular, the configuration $(D_{\mathrm{low}}, r)=(/, 192)$ exhibits a pronounced imbalance between the two dimensions of reduced projection matrices (\emph{i.e.} $\bm{U}$ and $\bm{V}$), which hinders adequate approximation of the high-dimensional FFN transformations. As a result, even though this configuration incurs higher computational cost for token selection, it shows substantially greater performance variation across datasets (\emph{e.g.}, achieving the best results on LSHT/LCC but the worst on Qasper/TQA), indicating reduced reliability of the proxy. 
These results suggest that, under a fixed selection-cost budget, a more balanced low-rank design can help improve modeling fidelity.

Accordingly, to achieve a practical trade-off between selection efficiency and prediction accuracy, we empirically adopt a moderate $(D_{\mathrm{low}}, r)$ configuration.

\begin{table*}[t]
\centering
\caption{Effect of selecting top-$\rho$ saliency values for channel importance on LLaMA. Results are reported on representative LongBench subsets and the average score (Avg.) on the full dataset. The default setting is underlined.}
{
\fontsize{8.6}{11}\selectfont
\setlength{\tabcolsep}{5pt}
\begin{tabular}{>{\centering}p{2.3cm}|ccccccccc|c}
\toprule
\multicolumn{1}{c|}{Method}       & MQA   & HPQA  & 2WiKi & MuSiQue & QMSum & TREC & LSHT & Count & LCC & Avg. (\%)\\
\midrule
$\rho=0.1$                 & 55.03 & 54.12 & 46.50 & 28.75 & 24.80 & 72.50 & 46.00 & 6.32 & 62.97 & 47.67  \\
\underline{$\rho=0.2$}     & 55.37 & 53.66 & 46.69 & 30.43 & 24.73 & 72.50 & 46.50 & 7.24 & 63.14 & 47.80 \\
$\rho=0.3$                 & 55.27 & 54.41 & 45.51 & 29.95 & 24.77 & 72.50 & 46.00 & 8.24 & 63.15 & \textbf{47.81} \\
w/o top-$\rho$             & 54.64 & 52.68 & 45.71 & 30.24 & 24.65 & 72.00 & 46.00 & 6.21 & 63.19 & 47.56 \\
\bottomrule
\end{tabular}
}
\label{tab:ablation_rho}
\end{table*}

\begin{table*}[t] 
    \centering
    \caption{Effects of starting layer (Left), stage number (Middle) and token budget (Right) on LLaMA, measured by LongBench average score (Avg.) and TTFT speedup ratio. The default setting is underlined.}
    \begin{minipage}[h!]{0.32\textwidth}
        \raggedright
        \setlength{\tabcolsep}{5pt}
        \fontsize{8.6}{11}\selectfont
        \begin{tabular}{c c c}
        \toprule[1pt]
        \makecell{Starting Layer} & \makecell{Avg. (\%)} & \makecell{TTFT Ratio}   \\
        \midrule
        8              & 47.43     & 1.78$\times$ \\
        \underline{10} & 47.80     & 1.68$\times$ \\
        12             & 48.01     & 1.60$\times$ \\
        \bottomrule[1pt]
        \end{tabular}
    \end{minipage}\hfill%
    \begin{minipage}[h!]{0.32\textwidth}
        \setlength{\tabcolsep}{5pt}
        \fontsize{8.6}{11}\selectfont
        \begin{tabular}{c c c}
        \toprule[1pt]
        \makecell{Stage Number}   & \makecell{Avg. (\%)}  & \makecell{TTFT Ratio}   \\
        \midrule
        3              & 47.62     & 1.66$\times$ \\
        \underline{4}  & 47.80     & 1.68$\times$ \\
        5              & 47.86     & 1.46$\times$ \\
        \bottomrule[1pt]
        \end{tabular}
    \end{minipage}\hfill%
    \begin{minipage}[h!]{0.32\textwidth}
        \raggedleft
        \setlength{\tabcolsep}{5pt}
        \fontsize{8.6}{11}\selectfont
        \begin{tabular}{c c c}
        \toprule[1pt]
        \makecell{Token Budget}   & \makecell{Avg. (\%)} &\makecell{TTFT Ratio}   \\
        \midrule
        (8K, 6K, 4K, 2K)               & 47.59     & 1.78$\times$ \\
        \underline{(9K, 7K, 4K, 2K)}   & 47.80     & 1.68$\times$ \\
        (10K, 8K, 4K, 2K)              & 47.79     & 1.61$\times$ \\
        \bottomrule[1pt]
        \end{tabular}
    \end{minipage}
    \label{tab:abaltion_starting_stage_budget}
\end{table*}

\begin{table*}[t]
\centering
\caption{Effect of the calibration data source for LTP on LLaMA. Results are reported on representative LongBench subsets and the overall average (Avg.) across the full dataset, compared with the second-best SlimInfer \cite{sliminfer}. The default setting is underlined.}
{
\fontsize{8.6}{11}\selectfont
\setlength{\tabcolsep}{5pt}
\begin{tabular}{>{\centering}p{2.3cm}|ccccccccc|c}
\toprule
\multicolumn{1}{c|}{Method}       & MQA   & HPQA  & 2WiKi & MuSiQue & QMSum & TREC & LSHT & Count & LCC & Avg. (\%)\\
\midrule
\makecell[l]{SlimInfer}                     & 53.72 & 54.34 & 43.68 & 30.69 & 24.60 & 71.00 & 45.50 & 6.40 & 63.38 & 47.56 \\
\makecell[l]{SPTS with C4}                  & 55.13 & 53.19 & 45.46 & 30.96 & 24.59 & 72.50 & 46.00 & 7.84 & 62.78 & 47.77 \\
\makecell[l]{\underline{SPTS with Qasper}}  & 55.37 & 53.66 & 46.69 & 30.43 & 24.73 & 72.50 & 46.50 & 7.24 & 63.14 & \textbf{47.80} \\
\bottomrule
\end{tabular}
}
\label{tab:ablation_calibration_set}
\end{table*}

\begin{table}[t]
    \centering
    \captionsetup{skip=1pt}
    \caption{Effect of the calibration set size on LLaMA, measured by LongBench Avg. score. The default setting is underlined.}
    \begin{minipage}[h!]{\linewidth}
        \centering
        \setlength{\tabcolsep}{5pt}
        \fontsize{8.6}{11}\selectfont
        \begin{tabular}{c | c c c c c c c}
        \toprule[1pt]
        Size     & 1     & 50    & 100   & 150   & \underline{200} & 250    & 300   \\
        Avg.(\%) & 47.52 & 47.68 & 47.73 & 47.70 & 47.80           & 47.82  & 47.80 \\
        \bottomrule[1pt]
        \end{tabular}
    \end{minipage}%
    \label{tab:ablation_calibration_size}
\end{table}

\subsection{Ablation on Query Token Length for PAP}
\label{sec:ablation_query_token_num}
Given that several prior works estimate token importance based on attention between input tokens and the last few tokens \cite{sliminfer, ftp}, we conduct an ablation on the length of query tokens used for probing partial attention in Eq.~(3). As reported in \cref{tab:ablation_window}, our default design, which relies solely the last token, achieves superior average performance.
Intuitively, under our token skipping strategy, which is primarily applied in deeper layers, the last token already provides strong task-oriented semantics and generalization capability. In contrast, averaging attention signals over multiple tokens may introduce semantic ambiguity in certain complex tasks, leading to noticeable performance fluctuations (\emph{e.g.}, on Count and TREC).
These findings further suggest a potential direction for future work: exploring task-aware adaptive probing strategies that align with diverse task characteristics, thereby facilitating more effective selective computation.

\subsection{Ablation on the Fusion Design for LTP}
\label{sec:ablation_fusion_type}
In order to identify tokens that simultaneously exhibit substantial feature transformations and task-relevant information, we construct a conditional selection criterion, as formulated in Eq.~(14). Intuitively, since the two criteria, $C_n^{\mathrm{FFN}}$ and $S_n^{\mathrm{MHA}}$, demonstrate distinct scales, we adopt a straightforward multiplicative fusion strategy, which avoids the need for explicit normalization and incurs minimal computational overhead. 

Nevertheless, we still explore several alternative designs for combining the two criteria. As shown in~\cref{tab:ablation_fusion_method}, we consider two scaling-based weighted fusion strategies, \emph{i.e.}, max normalization and standardization prior to additive fusion (denoted as MaxNorm-Add and Std-Add, respectively), yet neither yields competitive performance. 
We also investigate a sequential selection strategy, where subsets of active tokens are selected based on $S_n^{\mathrm{MHA}}$ and $C_n^{\mathrm{FFN}}$ in succession (denoted as Sequential Select). This also leads to unstable performance, with an average degradation of approximately 0.4\% compared to the default design.
We attribute these suboptimal performances to the implicit assignment of equal contribution to the two criteria. While improved performance could potentially be achieved by carefully tuning their relative weights, doing so would introduce additional hyperparameters and may compromise the stability across models. Therefore, we retain the original multiplicative fusion design.

\subsection{Ablation on Top-$\rho$ Saliency Values}
\label{sec:ablation_top_rho}
We evaluate the effect of utilizing the top-$\rho$ fraction of saliency values in Eq.~(10). As reported in \cref{tab:ablation_rho}, retaining only the top values for channel-wise importance estimation consistently improves the average accuracy, achieving up to a 0.25\% gain over using the full calibration set (\emph{i.e.}, w/o top-$\rho$). This design is motivated by the general observation that intermediate activations in LLMs exhibit significant variations \cite{systematic_outliers}. Therefore, intuitively, prioritizing the consistently highly activated dimensions preserves the most informative channels, enabling the proxy sub-network to better approximate the original FFN transformations and more accurately identify crucial tokens for reduced computation. Ultimately, we fix $\rho=0.2$ for most LLM-based experiments, which provides more consistent improvements across subsets and performs well across different long-context LLMs.

\subsection{Ablation on Stage Partitioning}
\label{sec:ablation_stage_partition}
We further ablate the configuration of stage partitioning, including the starting layer for token skipping, the stage number, and the per-stage token budgets, as reported in~\cref{tab:abaltion_starting_stage_budget}. Earlier starting layers and smaller token budgets yield higher acceleration but incur more substantial accuracy drops, while an appropriate number of stages better balances redundancy removal and information reuse. Notably, these hyperparameters entail an accuracy-efficiency trade-off and thus cannot be tuned independently under a fixed speedup budget. For simplicity and fairness, we adopt a fixed protocol in most cases: using the same starting layer as \cite{sliminfer,pyramiddrop}, 4 uniformly divided stages, and linear-like decreasing token budgets, with slight adjustments to match baseline speedups, as detailed in~\cref{sec:more_implementation_details}. 

Inevitably, different models exhibit distinct intrinsic behaviors and may require further budget-aware adjustments. Therefore for unseen models, we suggest keeping the first/last 3 layers unskipped, starting around the 1/3 layer for speedups $<2.5\times$, and earlier for higher speedups.

\subsection{Ablation on the Source of Calibration Data}
\label{sec:ablation_calibration_data}
In the LTP module, a data-driven parameter pruning scheme is adopted to evaluate and reduce the intermediate dimensions for the low-rank proxy network. For all LLM-oriented experiments, we follow prior work \cite{fitprune} by utilizing the fixed 200-sample Qasper subset as the calibration set. Since no ground-truth answers are accessed, this does not introduce data leakage from the evaluation set, rather, it may help reveal task-related parameter activation patterns and further facilitate accuracy preservation \cite{fitprune}. 
Despite this, several studies in parameter compression advocate the use of external datasets to estimate more stable and general parameter importance \cite{wanda, sparsegpt}. Motivated by this, we further investigate the generalization capability of SPTS under such settings. 
Concretely, we sample 200 sequences, each of length $2$K, from the C4 training set \cite{c4dataset}, consistent with the practice in \cite{sparsegpt,wanda}.

As shown in~\cref{tab:ablation_calibration_set}, SPTS achieves similar overall average scores across difference calibration data, and still outperforms the second-best method, SlimInfer \cite{sliminfer}. These results suggest that for FFNs, which are relatively task-agnostic modules, the prominently activated parameters tend to be stable. Therefore, the gains of our method primarily stem from this stable approximation, rather than from overfitting to the distribution of the evaluation datasets.

\textbf{Randomness.} Notably, using a broader calibration corpus introduces randomness through data sampling. We therefore further assess the sensitivity of SPTS to calibration sampling on C4 \cite{c4dataset}, following prior calibration-based methods \cite{sparsegpt,wanda}. Specifically, we randomly sample calibration sets across five seeds while keeping all inference settings fixed. On LLaMA, SPTS achieves a LongBench average score of 47.75±0.02\%, consistently outperforming the second-best result of 47.56\% achieved by SlimInfer \cite{sliminfer}. Across the same seeds, SPTS yields average score gains of 0.46±0.05\% and 1.58±0.09\% in the module-wise ablations reported in Tabs.~6 and~7, respectively, further confirming its robustness to randomness.

\subsection{Ablation on Calibration Set Size}
\label{sec:ablation_calibration_size}
Following \cite{sparsegpt,wanda}, we assess the sensitivity of SPTS to the calibration set size by varying the number of calibration samples drawn from LongBench, \cref{tab:ablation_calibration_size} shows. Since parameter activation patterns are statistical in nature, an overly small calibration set may fail to estimate them reliably, thereby degrading the accuracy of the proxy network. However, performance stabilizes once the calibration set exceeds 100 samples. To avoid sampling randomness and reduce calibration cost, we simply fix the 200-sample Qasper subset in all corresponding experiments.

\end{document}